\documentclass[runningheads]{llncs}
\usepackage{graphicx}

\usepackage{tikz}
\usepackage{comment}
\usepackage{amsmath,amssymb} 
\usepackage{color}
\usepackage{multirow}
\usepackage{booktabs}
\usepackage{makecell}
\usepackage{float}

\usepackage{cite}
\usepackage[pdftex, pagebackref=true,breaklinks=true,letterpaper=true,colorlinks,bookmarks=false, citecolor=green, linkcolor=red]{hyperref}
\usepackage{bm}

\usepackage[width=122mm,left=12mm,paperwidth=146mm,height=193mm,top=12mm,paperheight=217mm]{geometry}

\usepackage{xspace}
\makeatletter
\DeclareRobustCommand\onedot{\futurelet\@let@token\@onedot}
\def\@onedot{\ifx\@let@token.\else.\null\fi\xspace}
\def\eg{\emph{e.g}\onedot} 
\def\ie{\emph{i.e}\onedot}

\makeatother

\begin{document}
	\pagestyle{headings}
	\mainmatter
	\def\ECCVSubNumber{0000}  
	
	\title{Efficient and Degradation-Adaptive Network for Real-World Image Super-Resolution}
	
	\titlerunning{Degradation-Adaptive Super-Resolution (DASR)}
	%
	\author{\textbf{Jie Liang}$^{1}$,  \textbf{Hui Zeng}$^{2}$ and \textbf{Lei Zhang}$^{1}$\\
	}
	\authorrunning{J. Liang et al.}
	\institute{$^1$The HongKong Polytechnic University,  $^2$OPPO Research\\
		\textit{\{liang27jie,\, cshzeng\}@gmail.com}; \textit{cslzhang@comp.polyu.edu.hk}\\}
	
	\maketitle
	
	\renewcommand{\thefootnote}{\fnsymbol{footnote}}
	\footnotetext[1]{This work is supported by the Hong Kong RGC RIF grant (R5001-18).}
	
	\vspace{-1em}
	
	\begin{abstract}
		Efficient and effective real-world image super-resolution (Real-ISR) is a challenging task due to the unknown complex degradation of real-world images and the limited computation resources in practical applications. Recent research on Real-ISR has achieved significant progress by modeling the image degradation space; however, these methods largely rely on heavy backbone networks and they are inflexible to handle images of different degradation levels. In this paper, we propose an efficient and effective degradation-adaptive super-resolution (DASR) network, whose parameters are adaptively specified by estimating the degradation of each input image. Specifically, a tiny regression network is employed to predict the degradation parameters of the input image, while several convolutional experts with the same topology are jointly optimized to specify the network parameters via a non-linear mixture of experts. The joint optimization of multiple experts and the degradation-adaptive pipeline significantly extend the model capacity to handle degradations of various levels, while the inference remains efficient since only one adaptively specified network is used for super-resolving the input image. Our extensive experiments demonstrate that the proposed DASR is not only much more effective than existing methods on handling real-world images with different degradation levels but also efficient for easy deployment. Codes, models and datasets are available at \href{https://github.com/csjliang/DASR}{https://github.com/csjliang/DASR}.
	\end{abstract}
	
	\section{Introduction}
	
	Single image super-resolution (SISR)~\cite{johnson2016perceptual, wang2018esrgan, zhang2018image, ma2020structure, sun2010gradient} is an active research topic in low-level vision, aiming at reconstructing a high-resolution (HR) version of a degraded low-resolution (LR) image. Since the seminal work of SRCNN~\cite{dong2014learning}, many convolutional neural network (CNN) based SISR methods~\cite{sajjadi2017enhancenet, kim2016deeply, soh2019natural, zhang2018residual, jo2021tackling} have been proposed, most of which assume a pre-defined degradation process (\eg, bicubic down-sampling) from HR to LR  images. Despite the great success, the performance of these non-blind SISR methods will be much deteriorated when facing real-world images~\cite{liu2021blind} because of the mismatch of degradation models between the training data and the real-world test data~\cite{zhang2021designing}.
	
	The blind image super-resolution (BISR) methods~\cite{liu2021blind, zhang2018learning, luo2020unfolding, gu2019blind, zhou2019kernel} have been proposed to address the problems of non-blind SISR methods by considering more complex degradation kernels extracted from real-world images. However, the degradation space of these methods is actually restricted to a set of pre-collected kernels, such as the DPED kernel pool~\cite{zhou2019kernel, ignatov2017dslr}. For real-world images, their degradation space can be much larger, including more types and more complex kernels than the DPED kernel pool, more complex and stronger noise, and other degradation operations such as compression. Therefore, many recent researches have been focused on the real-world image super-resolution (Real-ISR) tasks~\cite{cai2019toward, wei2020component, lugmayr2020ntire, lugmayr2019aim, fritsche2019frequency, lugmayr2019unsupervised, ji2020real, ren2020real} by modeling and synthesizing the complex degradation process of real-world images~\cite{bulat2018learn, wei2021unsupervised}. The representative works include BSRGAN~\cite{zhang2021designing} and Real-ESRGAN~\cite{wang2021realesrgan}, which introduce comprehensive degradation operations such as blur, noise, down-sampling, and JPEG compression, and control the severity of each operation by randomly sampling the respective hyper-parameters. They also employ random shuffle of degradation orders~\cite{zhang2021designing} and second-order degradation~\cite{wang2021realesrgan} to better simulate the real-world complex degradations, respectively. 
	
	Despite the remarkable progress of BSRGAN~\cite{zhang2021designing} and Real-ESRGAN~\cite{wang2021realesrgan} on improving the image perceptual quality, they have several limitations for practical usage. On one hand, they are basically designed to work on severely degraded LR images. While BSRGAN and Real-ESRGAN can generate a certain amount of details on some tough LR images, they are difficult to generate fine details on mildly degraded LR inputs. It is highly anticipated to develop Real-ISR models which can handle images with different degradation levels. On the other hand, the BSRGAN and Real-ESRGAN methods rely on heavy backbone networks (\eg, RRDB~\cite{wang2018esrgan}), which make them difficult to be deployed on devices with limited computational resources~\cite{dong2016accelerating, zhang2021edge, ahn2018fast, yang2019lightweight, song2021addersr}. It is also anticipated to develop efficient Real-ISR models to meet the requirement of high efficiency.
	
	To tackle the above problems, in this paper, we propose a degradation-adaptive super-resolution (DASR) network whose parameters are adaptively specified to the given image according to its degradation. Our DASR consists of a tiny regression network to estimate the degradation parameters of the input image and multiple light-weight super-resolution experts, which are jointly optimized on a balanced degradation space. For each input image, an adaptive network is constructed via a non-linear mixture of experts, whose adaptive weighting factors are specified by the estimated degradation parameters. The multiple super-resolution experts and the degradation-aware mixture significantly improve the model capacity for handling images of different degradations. Meanwhile, the whole pipeline of DASR is highly efficient to meet the requirement of Real-ISR tasks, as only one adaptive network is employed to super-resolve the image during inference and the cost of mixing experts is negligible.
	
	The contributions of this paper are two-fold. First, we propose a degradation-adaptive super-resolution network, which significantly improves the model capacity to super-resolve images of various degradation levels. Second, the pipeline of our DASR network is highly efficient, providing a good solution to perform Real-ISR in practical applications. Extensive experiments verified the effectiveness and efficiency of the proposed method.
	
	\section{Related Work}
	
	\subsection{Real-World Image Super-Resolution}
	
	How to reproduce effectively and efficiently the HR image from low-quality and low-resolution real-world images is a challenging issue in SISR research. The distribution of real-world images can differ dramatically due to the varying image degradation process, different imaging devices, and image signal processing methods~\cite{liu2021blind, wei2021unsupervised}. Researches~\cite{cai2019toward, zhang2019zoom} have tried to capture real-world HR-LR image pairs by adapting the focal length of the camera, yet the collection of data is tedious and this can only describe a limited subspace of image degradation. There are also some unsupervised methods~\cite{wei2021unsupervised, fritsche2019frequency} proposed to explore the domain adaptation between the synthesized LR image and the real one, yet the domain gap is still inevitable which deteriorates the SR performance~\cite{lugmayr2019aim, lugmayr2020ntire}.
	
	Recently, several Real-ISR methods such as BSRGAN~\cite{zhang2021designing}, Real-ESRGAN~\cite{wang2021realesrgan} and SwinIR~\cite{liang2021swinir} have achieved remarkable progress by introducing comprehensive degradation models to effectively synthesize real-world images. However, they rely on a heavy and computationally intensive backbone network, \eg, RRDB~\cite{wang2018esrgan} and Swin transformer~\cite{liu2021swin}, and are not flexible to process images of different degradation levels. In this paper, we propose a degradation-adaptive framework to address this issue, targeting an effective and efficient network for the challenging Real-ISR task.
	
	\subsection{Image Degradation Modeling}
	
	In many non-blind SISR methods~\cite{ledig2017photo, wang2018esrgan, zhang2018image, wang2018recovering, johnson2016perceptual, ma2020structure, fuoli2021fourier}, the degradation model is simply assumed as bicubic down-sampling or blurred down-sampling with a Gaussian kernel. The performance of these non-blind methods can be dramatically undermined when applied to images with different degradations~\cite{liu2021blind}. As a remedy, SRMD~\cite{zhang2018learning}, UDVD~\cite{xu2020unified} and some other methods~\cite{zhang2020deep, zhang2019deep} extend the degradation space to cover more blur kernels and noise levels, and use the degradation map as additional input to perform conditional SISR. While these methods can handle multiple degradations with a single model, they rely on accurate degradation estimation, which itself is also a challenging task.
	
	A few blind SISR methods have been proposed for unknown degradation~\cite{wang2021unsupervised, hui2021learning, liu2020learning, maeda2020unpaired, bulat2018learn, yuan2018unsupervised, wei2021unsupervised}. In KMSR~\cite{zhou2019kernel}, a kernel pool is constructed from real photographs using generative adversarial network~\cite{goodfellow2014generative}, followed by synthesizing training pairs in a more realistic way.
	Some methods like IKC~\cite{gu2019blind} and VBSR~\cite{cornillere2019blind} incorporate a blur kernel estimator into the SISR framework, which can be adaptive to images degraded from different blur kernels~\cite{luo2020unfolding, kim2021koalanet}. However, most of the blind SISR methods are trained with a pre-collected kernel pool~\cite{zhou2019kernel, ignatov2017dslr}, and hence they are not really blind and can hardly be generalized to real-world images.
	
	Recent Real-ISR methods such as BSRGAN~\cite{zhang2021designing} and Real-ESRGAN~\cite{wang2021realesrgan} further extend the degradation modeling space by incorporating comprehensive degradation types with randomly sampled degradation parameters to enhance the variation. The larger degradation space helps the trained Real-ISR model to improve the perceptual quality of some tough LR inputs. However, the degradation parameter sampling in BSRGAN and Real-ESRGAN is unbalanced to train a flexible network, limiting the trained model in generating fine details, especially for inputs with mild degradations. In this work, we propose to balance the degradation space by partitioning it into three levels with balanced frequencies. Such balanced space facilitates the optimization of our degradation-adaptive model on different degradation levels and brings a better approximation to the real-world LR images.
	
	\subsection{Mixture of Experts and Dynamic Convolution}
	
	The mixture of experts (MoE,~\cite{jacobs1991adaptive, jordan1994hierarchical, gross2017hard, aljundi2017expert}) is a long-standing method that calculates the weighted sum of multiple expert networks to improve the performance. A trainable gating network is employed to compute the weight for activating each expert~\cite{maeda2020fast}, usually based on an explicit (\eg, labeled classes) or implicit (content clustering) partition of the data. In this paper, we calculate the adaptive weight of experts according to the degradation of the image for the Real-ISR tasks. Besides, instead of activating all experts and calculating the weighted sum of outputs as in previous MoE methods~\cite{wang2020blind}, we adaptively mix the network parameters, resulting in only one adapted network for inference. Such a pipeline is effective and efficient due to the increased non-linearity and the fast inference.
	
	Dynamic convolution~\cite{chen2020dynamic, li2021revisiting} or conditional convolution~\cite{li2021omni, yang2019condconv} aims to enhance the feature representation capacity by making the convolutional parameters sample-adaptive. Most of the existing methods optimize multiple sets of convolutional parameters and learn feature self-attention to linearly combine the parameters. However, this pipeline introduces many computations to obtain self-attention, causing a trade-off between efficiency and effectiveness. In this paper, we achieve the non-linear mixture of experts via an adapted conditional convolution, where the conditions are the degradation parameters and the weighting factors are calculated once for all layers to keep efficiency.
	
	\section{Methodology}
	
	\begin{figure*}[t]
		\centering
		\includegraphics[width=0.98\textwidth]{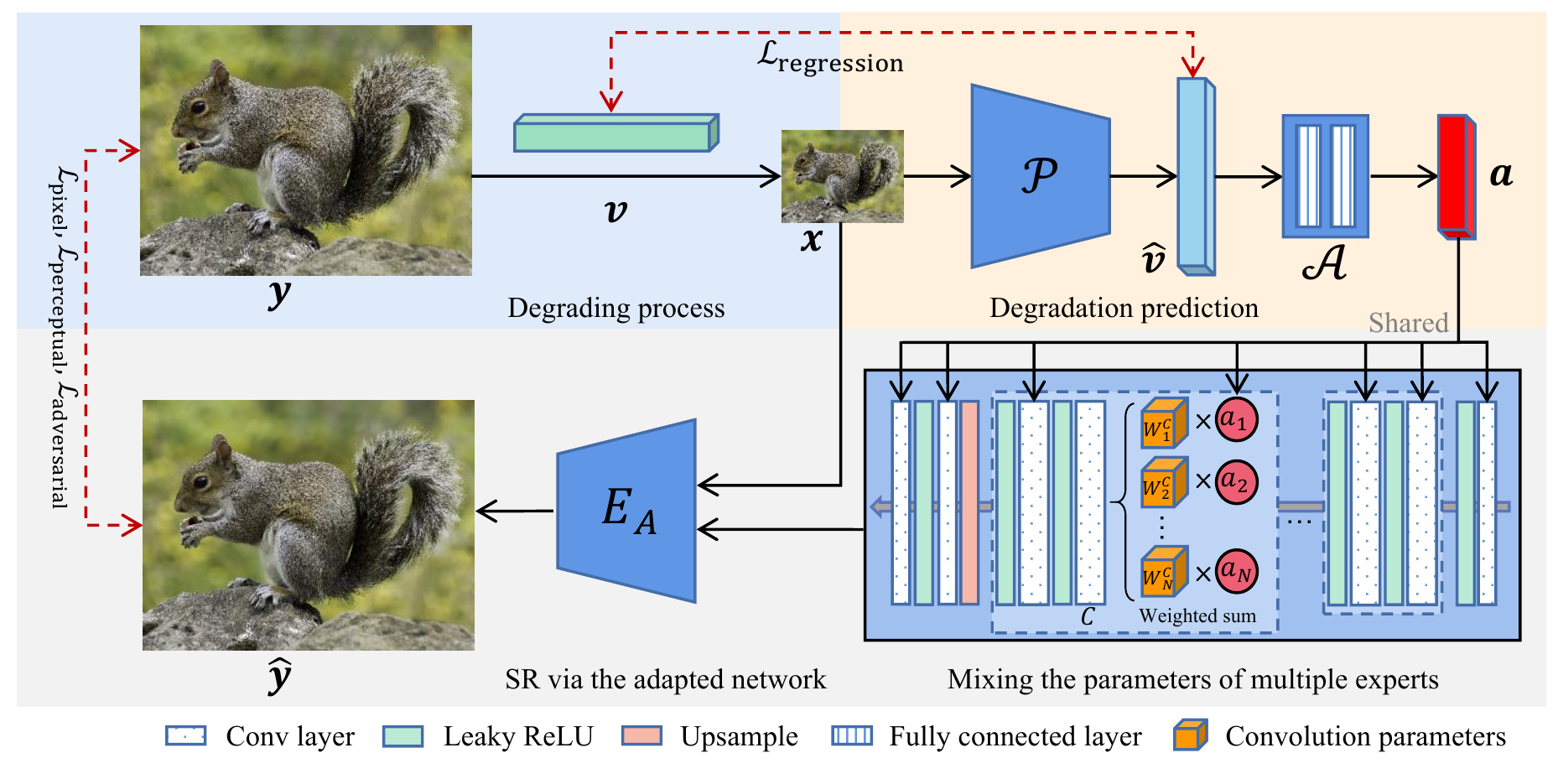}
		\vspace{-1em}
		\caption{\label{pipeline}Overall pipeline of the proposed DASR. Here, $\bm{x}, \bm{y}$ and $\hat{\bm{y}}$ denote the LR image, the ground truth HR image and the super-resolved result, respectively. For each convolution layer $C$, the parameters $W_i^C$ of $N$ experts are mixed according to the weighting factors in $\bm{a}$. The input $\bm{x}$ is super-resolved to $\hat{\bm{y}}$ by the adapted network $E_A$.}\vspace{-1em}
	\end{figure*}
	
	This section presents our degradation-adaptive network for real-world image super-resolution, \ie, DASR. As shown in Figure~\ref{pipeline}, DASR mainly consists of a degradation prediction network and a CNN-based SR network with multiple experts. In the following sections, we first provide the details of the proposed DASR framework and then introduce our degradation modeling to set degradation parameters and generate training pairs. 
	
	\subsection{Degradation-Adaptive Super-Resolution}
	\label{DASR}
	
	\textbf{Degradation prediction network.} To allow efficient and degradation-adaptive super-resolution, we propose to estimate the degradation parameters $\bm{v}\in\mathbb{R}^{1\times n}$ of each input $\bm{x}$ via a regression network $\mathcal{P}$, \ie, $\hat{\bm{v}} = \mathcal{P}(\bm{x})$, where $\hat{\bm{v}}$ denotes the estimation of $\bm{v}$. 
	We employ a set of parameters $\bm{v}$ to elaborately describe the degradation space. The details of degradation space modeling will be discussed in Section~\ref{degradation_model}. 
	To make the estimation process efficient, we design a light-weighted network $\mathcal{P}$ to predict $\bm{v}$. Specifically, $\mathcal{P}$ consists of $6$ convolution layers with Leaky ReLU activation, followed by a global average pooling layer. We first use convolution layers to extract image spatial degradation features and then use the global average pooling layer to estimate the degradation parameters. 
	
	To optimize the network $\mathcal{P}$, we introduce a regression loss between the estimated degradation parameters $\hat{\bm{v}}$ and the ground-truth $\bm{v}$ using the $\ell_1$-norm distance as follows:
	\begin{equation}
	\label{regression_loss}
	\mathcal{L}_{\text{regression}} = \lVert \hat{\bm{v}} - \bm{v} \rVert_1.
	\end{equation}
	According to the degradation model, each parameter in $\bm{v}$ is randomly sampled to specify the degradation process to generate the LR-HR image pairs.
	
	\textbf{Image super-resolution network.} An ideal Real-ISR method is expected to be both effective and efficient. On one hand, in real-world SR tasks, the computation resources are usually limited, especially for edge devices. On the other hand, the model should be able to effectively handle images with various kinds of degradations. Nevertheless, most of the current SR methods~\cite{liang2021swinir, zhang2021designing, wang2021realesrgan, ledig2017photo, lim2017enhanced} can only trade-off between efficiency and effectiveness, and they are inflexible to handle images with different degradation types and levels.
	
	To develop an effective and efficient Real-ISR model, we propose a degradation-adaptive SR network to boost the model capacity via a non-linear mixture of experts (MoE) strategy, whose additional cost is negligible during inference. In specific, we employ $N$ convolutional experts, denoted by $\bm{E} = [E_1, E_2, \cdots, E_N]$, where each expert $E_i$ is a light-weighted SR network, \eg, SRResNet~\cite{ledig2017photo} or EDSR-M~\cite{lim2017enhanced}, with independent parameters $\bm{\Phi}_{E_i}$. All the $E_i$ share the same network topology, and they are optimized jointly with the supervision of the same loss. Our idea is to implicitly train each expert to handle images falling into a sub-space of the degradation space so that they can work together to process images with various kinds of degradations in the whole space.
	
	A vector of weighting factors $\bm{a}\in\mathbb{R}^{1\times N}$, which is adaptive to the degradation of the input $\bm{x}$, is then calculated to adaptively mix the $N$ experts. We calculate $\bm{a}$ conditioned on the estimated $ \hat{\bm{v}} $ via a tiny network $\mathcal{A}$ with two fully-connected layers, \ie, $\bm{a} = \mathcal{A}(\hat{\bm{v}})$. As both $\hat{\bm{v}}$ and $\bm{a}$ are of low dimension ($n=33$ and $N=5$ in our experiments), the network $\mathcal{A}$ is highly efficient. Note that if $\bm{a}$ is constrained to be a one-hot vector, only one expert will be activated for super-resolving the input $\bm{x}$, and this will degrade our framework to a competitive MoE~\cite{maeda2020fast}, which may perform well on tasks whose sample distribution space can be partitioned with clear boundaries, yet it can hardly work well for the Real-ISR task with a large and continuous degradation space.
	
	With the multiple experts $\bm{E}$ and their adaptive weighting factors $\bm{a}$, we mix the experts adaptively in a non-linear manner. For each convolution layer $C$ of the desired network, we employ the dynamic convolution technique~\cite{yang2019condconv, chen2020dynamic} to parameterize the convolutional kernels as follows:
	\begin{equation}
	\bm{f}_{\text{output}} = \sigma((a_1\cdot W^C_1 + a_2\cdot W^C_2 + \cdots + a_N\cdot W^C_N)\ast \bm{f}_{\text{input}}).
	\end{equation}
	where $ \bm{f}_{\text{input}} $ and $ \bm{f}_{\text{output}} $ denote the input and the output features, $a_i$ indicates the $i^{th}$ value of $\bm{a}$, $W^C_i$ denotes the layer $C$ parameters for expert $E_i$ and $\sigma$ is the activation function. That is, we adaptively fuse the parameters of each layer among all experts, resulting in an adaptive network, denoted as $E_A$. 
	
	Note that in classic dynamic convolution, the weighting factor of each layer is calculated by an independent network conditioned on the feature map of the last layer, thus introducing non-negligible computational costs. In contrast, we learn a single set of degradation-adaptive weighting factors $\bm{a}$ for all convolution layers, which is very efficient. Our framework follows the spirit of MoE but in a non-linear manner due to the activation operation in intermediate layers. The non-linearity and the degradation-adaptive mixture of multiple experts significantly extend the model capacity to handle degradations of various levels.
	
	Our DASR is very efficient. For each convolutional layer, the model only deploys one adapted network $E_A$ in the inference stage, rather than deploying $N$ models as done in the classic MoE methods~\cite{jacobs1991adaptive, jordan1994hierarchical}. The degradation prediction network $\mathcal{P}$ and the weighting module $\mathcal{A}$ are also very light-weighted. Therefore, the cost of inference is of the same order as one single expert network. The computational overhead caused by the mixture operation is negligible. Specifically, the mixture process consists of multiplications and additions operations of the parameters of $N$ experts. For a light-weighted backbone network like SRResNet or EDSR-M, the number of parameters of each expert is only $1.52M$, and they are independent of the size of input images. Therefore, compared with the calculation of multiple feature maps, the complexity of the mixture of parameters is several orders of magnitude lower and thus can be neglected.
	
	\subsection{Degradation Modeling}
	\label{degradation_model}
	
	Since high-quality real-world LR-HR pairs are hard to be collected due to the misalignment issue~\cite{cai2019toward, zhang2019zoom}, the degradation modeling is very important to synthesize real-world LR inputs $\bm{x}$ from a given HR image $\bm{y}$ for Real-ISR model training. A degradation space, denoted by $S$, should be pre-defined to synthesize training pairs and perform degradation-adaptive optimization. The quality of an LR sample $\bm{x}$ in $S$ is controlled by a degradation parameter vector $\bm{v}=[v_1, v_2, \cdots, v_n]$, where $v_i$ specifies the type or severity of a degrading operation and $n$ denotes the number of degradation parameters. In our DASR, $\bm{v}$ also serves as the ground-truth for training the degradation prediction network.
	
	The image degradation model has been recently improved significantly from the simple bicubic down-sampling~\cite{dong2014learning, wang2018esrgan} to shuffling~\cite{zhang2021designing} and second-order~\cite{wang2021realesrgan} pipelines. We adopt the degradation operations of blurring (both isotropic and anisotropic Gaussian blur), resizing (both down-sampling and up-sampling with area, and bilinear and bicubic operations), noise corruption (both additive Gaussian and Poisson noise), and JPEG compression in our modeling. In $\bm{v}$, we use a one-hot code to quantify the degradation operation type and use a single value to record the degradation level normalized by its respective dynamic range. 
	
	It is worth mentioning that different from the methods~\cite{zhang2018learning, gu2019blind} which quantify a blur kernel by its kernel coefficients, we quantify a blurring degradation by its kernel size $s$, the standard deviation $\sigma_1, \sigma_2$ along the two principal axes, and the rotation degree $\theta$. In this way, the degradation parameters are more interpretable to specify the degradation types and levels, and can better support the degradation-aware mixture of experts. Meanwhile, the parameter vector $[s, \sigma_1, \sigma_2, \theta]$ has only $4$ dimensions, while the kernel vector $\bm{k}$ will have much higher dimensions to estimate. Benefiting from the interpretability and compactness of the degradation space, our DASR allows explicit user control towards degradation parameters during inference. This can facilitate many user-interactive applications to customize the desired super-resolving effect.
	
	Though the shuffling degradation method in BSRGAN~\cite{zhang2021designing} and the second-order degradation pipeline in Real-ESRGAN~\cite{wang2021realesrgan} can generate a sufficiently large degradation space, it is hard for them to train a model which can adaptively handle images with different levels of degradations. Our DASR is designed to be adaptive to a wide range of real-world inputs with multiple light-weight expert networks, each of which is expected to handle a subspace of images of different degradation levels. Therefore, we partition the whole degradation space $S$ into $3$ levels $[S_1, S_2, S_3]$ by specifying the parameters $\bm{v}$ accordingly. Among them, $S_1$ and $S_2$ are generated with first-order degradation with small and large parameter ranges, respectively, while $S_3$ is generated by the second-order degradation. Due to space limitation, more details of the degradation operations and the specification of $[S_1, S_2, S_3]$ are provided in the Section~\ref{Detail_degrade} in the Appendix.
	
	\subsection{Training Losses}
	\label{train_loss_section}
	
	The learnable modules of our DASR network include $[\bm{E}, \mathcal{P}, \mathcal{A}]$. As mentioned in Section 3.1, the $\mathcal{L}_{\text{regression}}$ loss is used to optimize $\mathcal{P}$ to predict the degradation parameters. To optimize the overall framework, following the many works in literature~\cite{wang2021realesrgan, zhang2021designing, wang2018esrgan}, we adopt the $L_1$-norm pixel-wise loss $\mathcal{L}_{\text{pixel}}$, the perceptual loss $\mathcal{L}_{\text{perceptual}}$ and the adversarial loss $\mathcal{L}_{\text{adversarial}}$. The total loss is defined as follows (more details are provided in the Section~\ref{Detail_loss} in the Appendix):
	\begin{equation}
	\label{train_loss_equation}
	\mathcal{L}_{\text{total}} = \mathcal{L}_{\text{pixel}} + \lambda_1 \mathcal{L}_{\text{regression}} + \lambda_2 \mathcal{L}_{\text{perceptual}} + \lambda_3 \mathcal{L}_{\text{adversarial}},
	\end{equation}
	where $\lambda_1, \lambda_2$ and $\lambda_3$ denote the balancing parameters.
	
	\section{Experiments}
	
	\subsection{Training Details}
	
	Following previous works~\cite{wang2018esrgan, wang2021realesrgan}, we employ DIV2K, Flickr2K, and OutdoorSceneTraining datasets for training our DASR model. For efficiency, we employ the SRResNet~\cite{ledig2017photo} as our backbone. The weights of the $N$ experts are initialized by the model pre-trained with pixel-wise loss. The Adam~\cite{kingma2014adam} optimizer is employed to train the network. The learning rate is set to $1\times e^{-4}$, the total batch size is $24$ and the training iteration is set to $500K$. We balance the training loss with $\lambda_1:\lambda_2:\lambda_3 = 1:1:0.1$. Without loss of generality and for a fair comparison, we conduct Real-ISR experiments with the scale factor of $4$ by following the setting in BSRGAN~\cite{zhang2021designing} and Real-ESRGAN~\cite{wang2021realesrgan}. In our experiment, the dimension of degradation parameters is $n=33$ and the number of experts is $ N=5$. The LR patch size is set to $64\times 64$.
	
	\subsection{Evaluation and Compared Methods}
	
	We evaluate our DASR method both quantitatively and qualitatively. For quantitative evaluation, as in BSRGAN~\cite{zhang2021designing} we synthesize $300$ LR-HR pairs by applying the $3$ levels of degradations to the $100$ validation images in the DIV2K dataset, \ie, $100$ LR-HR pairs for each level. We also make the comparison on the original DIV2K dataset with bicubic downsampling. An illustration of images with different degradations is shown in Fig.~\ref{levels}, where more samples are shown in Section~\ref{More_sample} in the Appendix. For qualitative evaluation, we also employ the images in the RealSRSet~\cite{zhang2021designing, wang2021realesrgan}, where the input images are corrupted by various blur, noise, or other real degradation operations.
	
	We compare the proposed DASR with representative and state-of-the-art SR methods, including RRDB~\cite{wang2018esrgan}, ESRGAN~\cite{wang2018esrgan}, IKC~\cite{gu2019blind}, BSRGAN~\cite{zhang2021designing}, Real-ESRGAN~\cite{wang2021realesrgan} and Real-SwinIR (-M and -L)~\cite{liang2021swinir}. Among them, RRDB is trained on bicubic degradation with pixel-wise loss; ESRGAN is trained on bicubic degradation with pixel-wise, perceptual and adversarial losses; IKC is a representative BISR method trained on various isotropic Gaussian blur kernels; BSRGAN and Real-ESRGAN are state-of-the-art Real-ISR methods with a heavy RRDB backbone; Real-SwinIR is trained on the degradation space of BSRGAN with the computationally expensive SwinIR backbone. 
	
	For a more comprehensive and fair comparison, we also re-train those commonly used backbone networks, including SRResNet, EDSR, RRDB, and SwinIR, with our constructed training dataset. Following the common practice~\cite{zhang2021designing, wang2021realesrgan}, we employ PSNR (the larger the better) and LPIPS (learned perceptual image patch similarity, the smaller the better) to quantitatively compare the performance of different methods on synthetic datasets, and make visual comparisons on real-world images since there are no reference images.
	
	\vspace{-1em}
	\subsection{Quantitative Comparison}
	\label{Quantitative_sec}
	
	\begin{figure*}[t]
		\centering
		\includegraphics[width=0.95\textwidth]{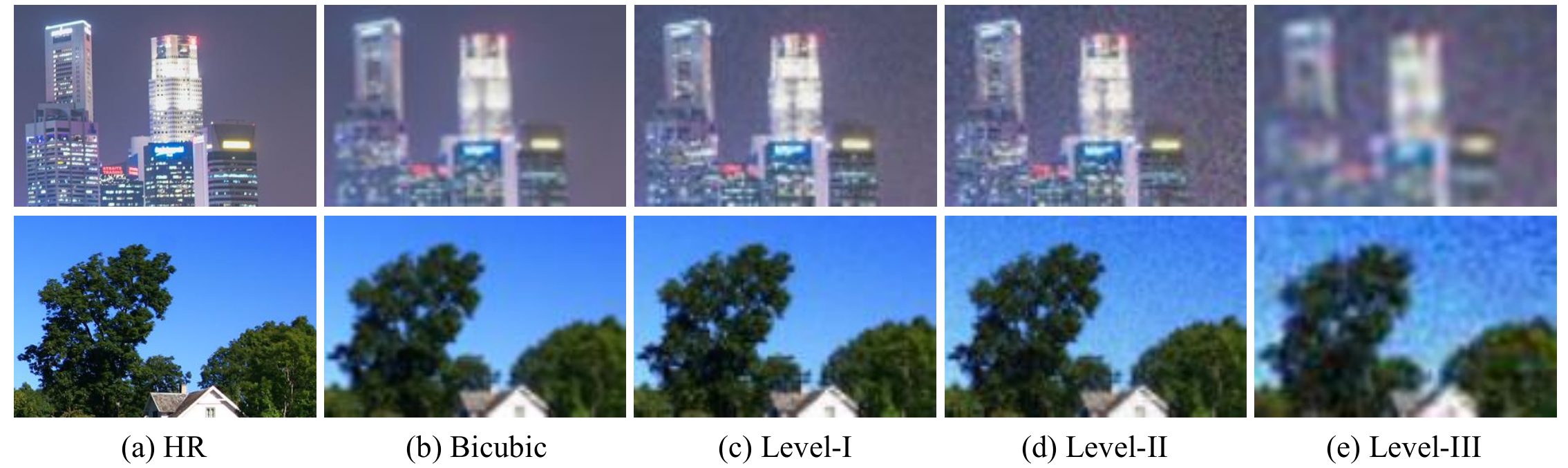}
		\vspace{-1em}
		\caption{\label{levels}Sample images with different levels of degradations in our datasets.}\vspace{-1.5em}
	\end{figure*}
	
	\textbf{Effectiveness.} In Table~\ref{effectiveness_ori} and Table~\ref{effectiveness_new}, we quantitatively compare the performance of competing methods in terms of PSNR and LPIPS on datasets with different levels of degradations. Specifically, Table~\ref{effectiveness_ori} compares the methods trained with their own degradation models, while Table~\ref{effectiveness_new} compares the methods re-trained on our proposed degradation space. 
	
	As shown in Table~\ref{effectiveness_ori}, existing methods can only achieve satisfactory performance on datasets with a specific type of degradation. For example, RRDB and ESRGAN can respectively achieve good fidelity and perceptual quality on the bicubic-downsampled dataset, yet their performance drops dramatically when handling images with other degradations, even for the `Level-I' degradation with mild noise and blurs. Real-ESRGAN, BSRGAN, and Real-SwinIR perform well on the most severely degraded dataset, \ie, `Level-III'. However, their performance deteriorates much on the other three datasets. 
	
	In contrast, our DASR achieves stable and significant improvement against other methods under the first three types of degradations, which cover the majority of real-world images, while achieving highly competitive (among the best two) results for the last type of degradation. For example, DASR outperforms Real-ESRGAN by about $1.7$dB in PSNR and $26\%$ in LPIPS on the `Level-I' dataset. On the `Level-III' dataset with severely degraded images (as shown in Fig.~\ref{levels} (d)), DASR achieves almost the same PSNR and LPIPS indices as BSRGAN. These observations clearly demonstrate that our DASR can generalize well to images with a wide range of degradations. 
	
	To further validate the effectiveness of our degradation-adaptive strategy, in Table~\ref{effectiveness_new} we re-train the backbones of popular SR models on our proposed degradation space. Note that the heavy RRDB backbone is adopted in both BSRGAN and RealESRGAN, and the lightweight SRResNet is adopted in our DASR as the backbone. As can be seen from this table, with the same network topology and similar computational overhead, our DASR outperforms the baseline SRResNet on all datasets by a large margin, \eg, improving $0.5$db of PSNR on the bicubic-downsampled dataset and about $5\%$ of LPIPS on the Level-II dataset. This demonstrates that the degradation-adaptive mixture of multiple experts can significantly extend the model capacity while keeping the efficiency.
	
	Compared to RRDB and SwinIR backbones that are adopted in recent state-of-the-art methods~\cite{zhang2021designing, wang2021realesrgan, liang2021swinir}, our DASR consumes much less computational resources, \eg, about $1/3$ and $1/12$ latency of RRDB and SwinIR, respectively. At the same time, DASR outperforms these heavy models in terms of reconstruction fidelity on all datasets, demonstrating its effectiveness of degradation-adaptive super-resolution and high efficiency to deploy in practice.

	\begin{table}[t]
		\caption{Quantitative comparison of different methods on datasets with different degradations (D-Level). `Bicubic' denotes the DIV2K validation set with bicubic degradation, while `Level I', `II', and `III' denote the datasets with mild, medium, and severe degradations, respectively. For the compared methods, we employ their officially released pre-trained models. The PSNR is calculated on the Y channel of YCbCr space.}
		\scriptsize
		\label{effectiveness_ori}
		\vspace{-2.5em}
		\begin{center}
			\begin{tabular}{p{1cm}p{1cm}<{\centering}p{1cm}<{\centering}p{1.2cm}<{\centering}p{0.8cm}<{\centering}p{1.3cm}<{\centering}p{1.2cm}<{\centering}p{1.4cm}<{\centering}p{1.4cm}<{\centering}p{1cm}<{\centering}}
				\toprule
				D-Level&Metric&RRDB&ESRGAN&IKC&BSRGAN&\makecell[c]{Real-\\ESRGAN}&\makecell[c]{Real-\\SwinIR-M}&\makecell[c]{Real-\\SwinIR-L}&DASR\\
				\midrule
				\multirow{2}{*}{Bicubic}&PSNR&30.92&28.17&28.01&27.32&26.65&26.83&27.21&28.55\\
				&LPIPS&0.2537&0.1154&0.2695&0.2364&0.2284&0.2221&0.2135&0.1696\\
				\midrule
				\multirow{2}{*}{Level-I}&PSNR&26.27&21.16&24.09&26.78&26.17&26.21&26.45&27.84\\
				&LPIPS&0.3419&0.4727&0.3805&0.2412&0.2312&0.2247&0.2161&0.1707\\
				\midrule
				\multirow{2}{*}{Level-II}&PSNR&26.46&22.77&25.39&26.75&26.16&26.12&26.39&27.58\\
				&LPIPS&0.4441&0.4900&0.4531&0.2462&0.2391&0.2313&0.2213&0.2126\\
				\midrule
				\multirow{2}{*}{Level-III}&PSNR&23.91&23.63&22.91&24.05&23.81&23.34&23.46&23.93\\
				&LPIPS&0.7631&0.7314&0.7583&0.3995&0.3901&0.3844&0.3765&0.4144\\
				\bottomrule
			\end{tabular}
		\end{center}\vspace{-2em}
	\end{table}

	\textbf{Efficiency.} The inference efficiency is a crucial factor in Real-ISR tasks due to the limited computational resources in practical applications. We compare different backbone networks in terms of multiple efficiency-related metrics and depict the results in the bottom rows of Table~\ref{effectiveness_new}.
	
	\begin{table}[t]
		\caption{Quantitative comparison of different backbone networks re-trained on our proposed degradation space and the efficiency comparison (the bottom rows). The evaluation datasets are the same as in Table~\ref{effectiveness_ori}. For efficiency evaluation, the input-dependent metric FLOPs is calculated on images with $256\times 256$ pixels; the Latency and Memory are the average inference time and the maximum GPU memory allocation on the DIV2K validation dataset (most LR inputs are with $510\times 339$ pixels). Statistics are collected following the implementation of~\cite{zhang2020aim, zhang2019aim} by using an NVIDIA V100 GPU.}
		\scriptsize
		\label{effectiveness_new}
		\vspace{-2em}
		\begin{center}
			\begin{tabular}{p{1.2cm}p{1.2cm}<{\centering}p{1.7cm}<{\centering}p{1.7cm}<{\centering}p{1.7cm}<{\centering}p{1.7cm}<{\centering}p{1.7cm}<{\centering}}
				\toprule
				\multicolumn{2}{c}{Data \& Metrics}&SRResNet&EDSR&SwinIR&RRDB&DASR\\
				
				\midrule
				\multirow{2}{*}{Bicubic}&PSNR&28.05&28.26&28.28&27.92&28.55\\
				&LPIPS&0.1747&0.1807&0.1488&0.1473&0.1696\\
				\midrule
				\multirow{2}{*}{Level-I}&PSNR&27.60&27.79&27.78&27.84&27.84\\
				&LPIPS&0.1772&0.1834&0.1531&0.1569&0.1707\\
				\midrule
				\multirow{2}{*}{Level-II}&PSNR&27.34&27.53&27.45&27.29&27.58\\
				&LPIPS&0.2228&0.2284&0.1854&0.1886&0.2126\\
				\midrule
				\multirow{2}{*}{Level-III}&PSNR&23.71&23.87&23.60&23.54&23.93\\
				&LPIPS&0.4419&0.4351&0.3869&0.3847&0.4144\\
				\midrule
				\multicolumn{2}{l}{Latency (ms)}&113&105&1719&460&142\\
				\multicolumn{2}{l}{\#FLOPs (GMac)}&166&130&539&1176&184\\
				\multicolumn{2}{l}{\#Params (M)}&1.52&1.52&11.72&16.70&8.07\\
				\multicolumn{2}{l}{\#Memory (M)}&2359&2169&2699&2417&2452\\
				\bottomrule
			\end{tabular}
		\end{center}\vspace{-2.5em}
	\end{table}
	
	As shown in the table, the computational overhead of different backbone networks differs dramatically. For example, RRDB~\cite{wang2018esrgan}, which is employed in recent Real-ISR methods~\cite{zhang2021designing, wang2021realesrgan}, consumes about $7$ times the FLOPs and more than $4$ times the inference time than SRResNet~\cite{ledig2017photo}. In other words, the RRDB based Real-ISR methods achieve superior performance at the price of applicability. The recent transformer-based method SwinIR has an acceptable number of FLOPs, however, it actually consumes much more inference time due to the heavy computation of attentions and frequent IO consumption.
	
	Benefiting from the light SRResNet-based backbone and the efficient degradation prediction and parameter fusion, our DASR is very efficient. In specific, the degradation prediction network $\mathcal{P}$ and the weighting module $\mathcal{A}$ consume $18$GMac FLOPs, $18$ms latency, $0.47$M parameters and $111$M GPU memory in total for $N=5$. Besides, the consumption on parameter fusion operation is negligible, as there are only $N\times 1.52$M multiplications and additions respectively and they can be calculated in parallel. Compared with the classical MoE methods that mix the feature maps of all experts~\cite{jacobs1991adaptive, jordan1994hierarchical, emad2022moesr, wang2020blind}, our DASR only conducts one forward pass. As a result, the computational cost increases slightly with a larger $N$, which supports a flexible extension of model capacity. 
	
	It is worth mentioning that although our model has more parameters, the maximum GPU memory consumption does not increase much as shown in the row of \#Memory in Table~\ref{effectiveness_new}, since the deployment of model parameters costs much less space than storing input-dependent feature maps. On the other hand, the increased model parameters do not demand much storage space, which is much easier to afford than the computing power. 
	
	\vspace{-1em}
	\subsection{Qualitative Comparison}
	
	\begin{figure*}[t]
		\centering
		\includegraphics[width=0.91\textwidth]{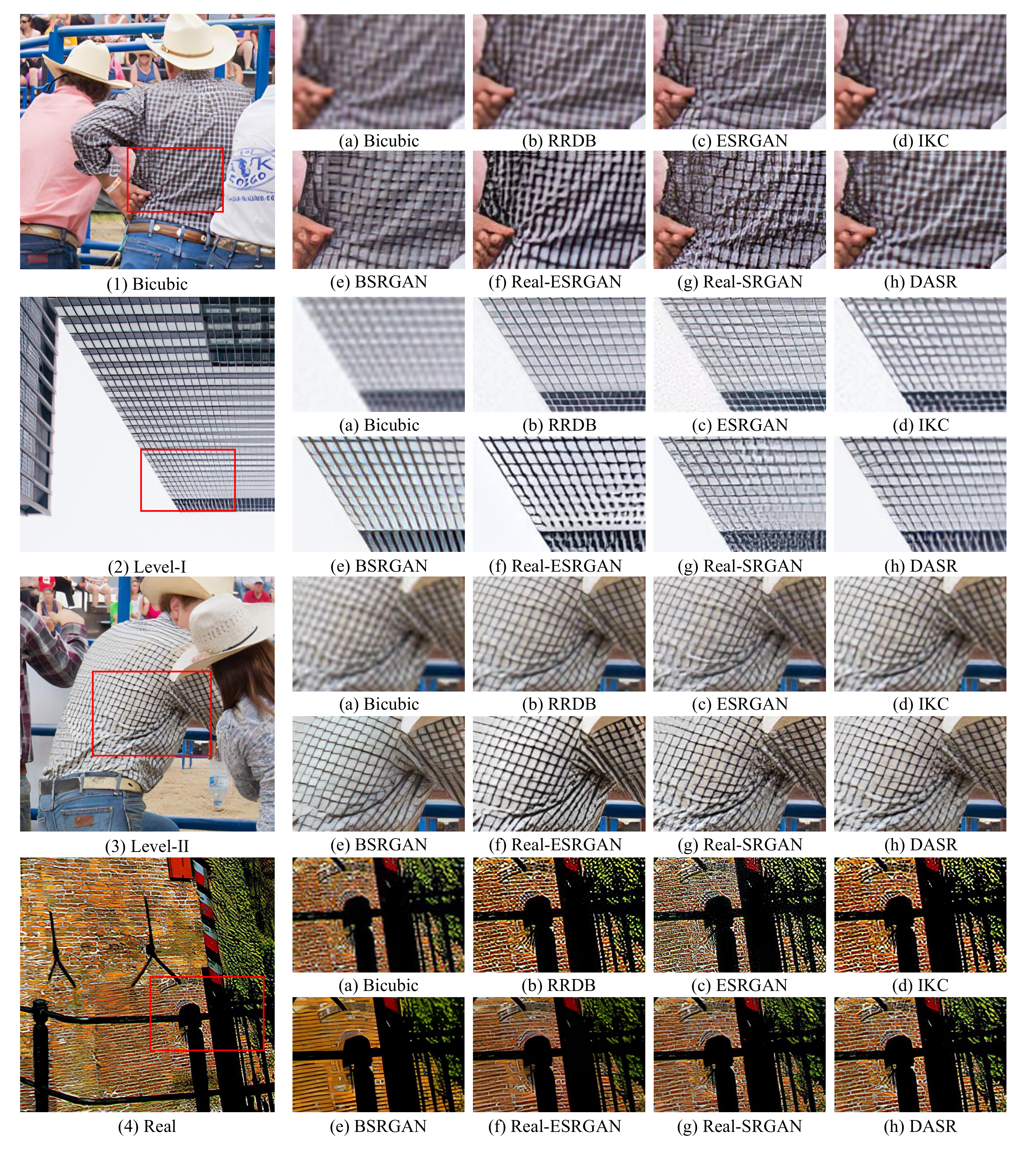}
		\vspace{-2.5em}
		\caption{\label{overallvisualcomp}Qualitative comparison of competing methods on images with different degradations. The results of (b-f) are generated by using the officially released models, while the output of (g) is obtained by re-training the SRResNet backbone with our proposed degradation model. Better zoom in for details.}\vspace{-1.5em}
	\end{figure*}
	
	Fig.~\ref{overallvisualcomp} shows the visual comparisons between different methods on images with different degradations. One can see that DASR can stably restore sharp and realistic details and remove artifacts for a wide range of degradations. In specific, the first sample image is degraded with bicubic downsampling and suffers from the aliasing issue. Both BSRGAN and Real-ESRGAN cannot generate satisfactory texture details even with the heavy RRDB backbone. This is because these two methods are trained on pairs with relatively severe degradations so that their denoising capacity is strengthened yet the detail-generation capacity is limited. Similar observations can be made on all the four samples in Fig.~\ref{overallvisualcomp}.
	
	The RRDB backbone trained with pixel-wise loss performs well on the first two samples in generating textures details, yet it cannot be generalized to the last two samples whose degradations are severe. This is reasonable since all its training pairs are generated by bicubic downsampling. In addition, the results of RRDB in the first and third samples are blurry, which is a well-acknowledged side-effect of pixel-wise loss. By applying perceptual and adversarial losses, ESRGAN achieves sharper results yet introduces many visual artifacts due to the instability of training generative adversarial networks. The ESRGAN also amplifies the noise as shown in the second sample. By considering different blur kernels, IKC can restore rich textures on most images, yet bring overshoot artifacts when facing unseen kernels in real-world images (the fourth sample). It also lacks the capacity to remove noise as shown in the second sample.
	
	The results of Real-SRGAN are obtained by re-training the SRResNet on our proposed degradation space with the same loss as in Real-ESRGAN~\cite{wang2021realesrgan}. It can be observed that due to the insufficient feature representation capacity, Real-SRGAN cannot perform well on all four samples compared to our DASR. In the first three samples, the Real-SRGAN generates messy details or artifacts, as the light-weighted model limits its capacity to achieve degradation-adaptive super-resolution. On the last sample which is a real-world image, the reconstruction of rich details is restricted in Real-SRGAN. In contrast, our proposed DASR outperforms the others in reconstructing realistic details and inhibiting artifacts, thanks to the effective degradation-adaptive framework and the joint optimization of multiple experts. More visual comparisons can be found in Section~\ref{More_comparison} in the Appendix.
	
	\begin{figure*}[t]
		\centering
		\includegraphics[width=0.93\textwidth]{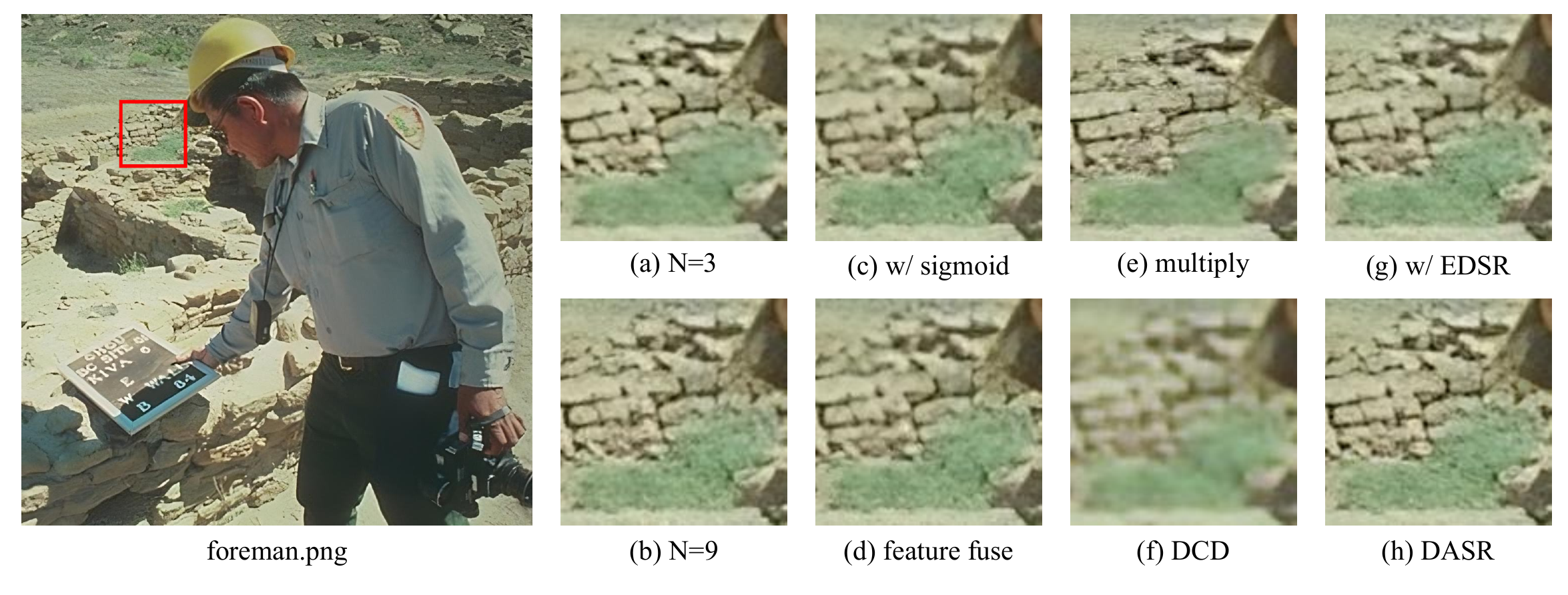}
		\vspace{-1.5em}
		\caption{\label{ablation}Ablation study. (a) and (b) validate the models with different $N$; (c) appends a sigmoid layer to the weighting module $\mathcal{A}$; (d) conducts classical MoE~\cite{jacobs1991adaptive, jordan1994hierarchical, emad2022moesr, wang2020blind} where the output of multiple experts are fused; (e) performs dynamic convolution with a single expert by learning a mapping matrix and multiplying it to the parameters; (f) conducts dynamic convolution following the work~\cite{li2021revisiting}; (g) applies EDSR-M backbone to DASR; (h) denotes our default DASR model.}\vspace{-1.2em}
	\end{figure*}
	
	\subsection{Ablation Study}
	
	We conduct comprehensive ablation studies on our proposed DASR model by using real-world images and depict the visual results in Fig.~\ref{ablation}. 
	
	\textbf{Effectiveness of $N$.} Models in Figs.~\ref{ablation}(a) and (b) evaluate the selection of $N$. It can be seen that using $3$ experts leads to relatively smooth results, while models of $N=5$ in (h) and $N=9$ in (b) enhance the generation of details. As $N=9$ shows similar visual quality to $N=5$, we consider that $N=5$ is sufficient to model the proposed degradation space. 
	
	\textbf{Effectiveness of model design.} Figs.~\ref{ablation}(c) and (d) validate the effectiveness of our model design. The result in (c) demonstrates that adding a sigmoid layer to the weighting module $\mathcal{A}$ cannot improve the performance. As we mix different experts in terms of model parameters, there is no need to ensure positive weights by a sigmoid layer. The experts in Fig.~\ref{ablation}(d) are fused following the strategy of classical MoE~\cite{jacobs1991adaptive, jordan1994hierarchical, emad2022moesr, wang2020blind}, where all experts are forwarded and the outputs are fused. We can see that the result of classical MoE in (d) lacks fine details compared to (h), yet its computational cost is $N$ times heavier than our DASR. 
	
	\textbf{Effectiveness of different dynamic convolutions.} Figs.~\ref{ablation}(e) and (f) compare different dynamic convolutions~\cite{li2021revisiting, xu2020unified} without introducing many additional parameters. While the inference latency and FLOPs are increased, the performance of those methods drops, \eg, the artifacts generated in (e). We believe it is the joint optimization of multiple experts and the degradation-adaptive mixture that make our DASR more effective than other methods.
	
	\textbf{Generalization to different backbone.} Fig.~\ref{ablation}(g) applies the EDSR-M backbone to DASR. The satisfactory perceptual quality of (g) demonstrates the generalization capacity of our proposed DASR to different backbone networks.
	
	\vspace{-0.5em}
	\subsection{User-Interactive Super-resolution}
	
	\begin{figure*}[t]
		\centering
		\includegraphics[width=0.88\textwidth]{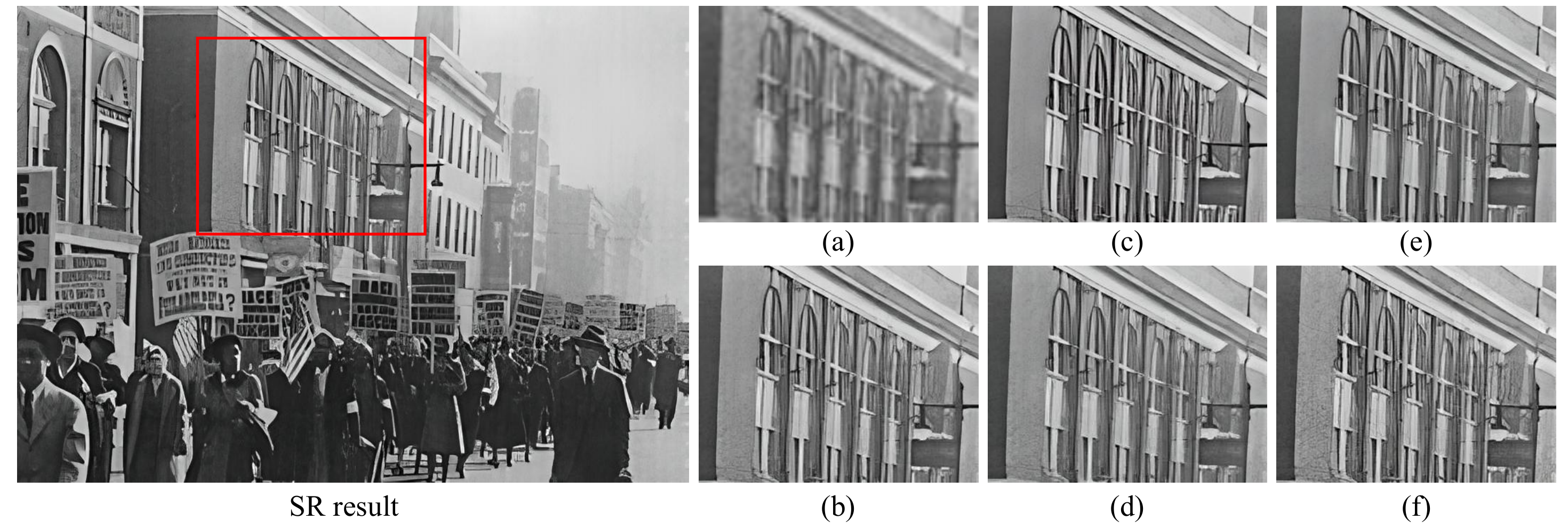}
		\vspace{-1em}
		\caption{\label{usercontrol}Example of user-interactive super-resolution. (a) is the input image with bicubic upsampling; (b) is the result of DASR where the degradation parameters are estimated automatically by model $\mathcal{P}$; (c) and (d) are generated by manually increasing and decreasing the scale of blur kernel, respectively; (e) and (f) are the super-resolution results by manually increasing and decreasing the level of noise, respectively.}\vspace{-1.2em}
	\end{figure*}
	
	One interesting advantage of our DASR over other Real-ISR methods is that it supports easy user-interactive super-resolution during inference, owing to its interpretable and compact degradation representation.
	
	We depict an example of user-interactive super-resolution in Fig.~\ref{usercontrol}. As can be seen, the proposed DASR allows explicit user control to customize the super-resolution effects. Manually setting larger values to the blur-related parameters (\eg, kernel scale) leads to sharper super-resolution results, as shown in Fig.~\ref{usercontrol}(c), while adjusting the level of noise can flexibly balance between image details and noise, as shown in Fig.~\ref{usercontrol}(e) and (f).
	Such an advantage of flexible user control makes our DASR very attractive in practical Real-ISR tasks.
	
	\section{Conclusion} 
	
	In this paper, we proposed an efficient degradation-adaptive network, namely DASR, for the real-world image super-resolution (Real-ISR) task. In order to improve the modeling capacity and flexibility of various degradation levels, we jointly learned multiple super-resolution experts and adaptively mixed them into one expert in a degradation-aware manner. The proposed DASR was not only degradation adaptive but also efficient during inference. Extensive quantitative and qualitative experiments were conducted. The results demonstrated that DASR not only achieved superior performance on images with a wide range of degradation levels but also kept good efficiency for easy deployment. In addition, DASR allowed easy user control for customized super-resolution results.
	
	\section{Appendix}

	\subsection{Detailed Settings of Degradation Modeling}
	\label{Detail_degrade}
	
	We report the detailed parameter settings of our degradation modeling in Table~\ref{parameter}.	
	We partition the whole degradation space $S$ into $3$ levels $[S_1, S_2, S_3]$, and randomly select one of them to generate the LR-HR image pairs during training with a balanced probability of $[0.3, 0.3, 0.4]$. For the blur operation, we use isotropic and anisotropic Gaussian kernels with a probability of $[0.65, 0.35]$, where we set $\sigma_1 = \sigma_2$ if isotropic blur kernel is specified. In the second degradation stage of $S_3$, following the practice in Real-ESRGAN~\cite{wang2021realesrgan}, we skip the blur operation with a probability of $0.2$, and perform sinc kernel filtering with a probability of $0.8$. We finally resize the image to the desired LR size, \ie, $1/4$ of the original size.
	
	For those operations that have more than one mode, \eg, the resize mode, we use a one-hot vector to indicate the choice of mode in $\bm{v}$. For other parameters, we normalize each of them by $v^\prime = (v - v_{\text{min}}) / (v_{\text{max}} - v_{\text{min}})$, where $v, v^\prime, v_{\text{min}}$ and $v_{\text{max}}$ indicate the original value, the normalized value, the minimum and maximum values of the parameter, respectively.

	\subsection{Details of Training Losses}
	\label{Detail_loss}
	
	As discussed in Section~\ref{train_loss_section}, the total training loss is defined as
	$$
	\mathcal{L}_{\text{total}} = \mathcal{L}_{\text{pixel}} + \lambda_1 \mathcal{L}_{\text{regression}} + \lambda_2 \mathcal{L}_{\text{perceptual}} + \lambda_3 \mathcal{L}_{\text{adversarial}},
	$$
	where the regression loss $\mathcal{L}_{\text{regression}}$ has been provided in Eq.~\eqref{regression_loss} of the main paper. For the other three losses, the settings are the same as in Real-ESRGAN~\cite{wang2021realesrgan}. Specifically, the pixel loss is defined as the $\ell_1$ distance $\mathcal{L}_{\text{pixel}} = \lVert \hat{\bm{y}} - \bm{y} \rVert_1$, where $\hat{\bm{y}}$ and $\bm{y}$ denote the super-resolved image and the ground-truth HR image, respectively. For the perceptual loss $\mathcal{L}_{\text{perceptual}}$, we first extract the \{conv$_1$, conv$_2$, conv$_3$, conv$_4$, conv$_5$\} feature maps of $\hat{\bm{y}}$ and $\bm{y}$ by using the pre-trained VGG19 network~\cite{simonyan2014very}, then calculate the weighted sum of the respective $\ell_1$ distances between the feature maps of $\hat{\bm{y}}$ and $\bm{y}$ as the perceptual loss, where the weights are set to be [0.1, 0.1, 1, 1, 1]. For the adversarial loss $\mathcal{L}_{\text{adversarial}}$, the U-Net discriminator with spectral normalization is adopted.
	
	\subsection{More Sample Images}
	\label{More_sample}
	
\begin{figure*}[t]
		\centering
		\includegraphics[width=1\textwidth]{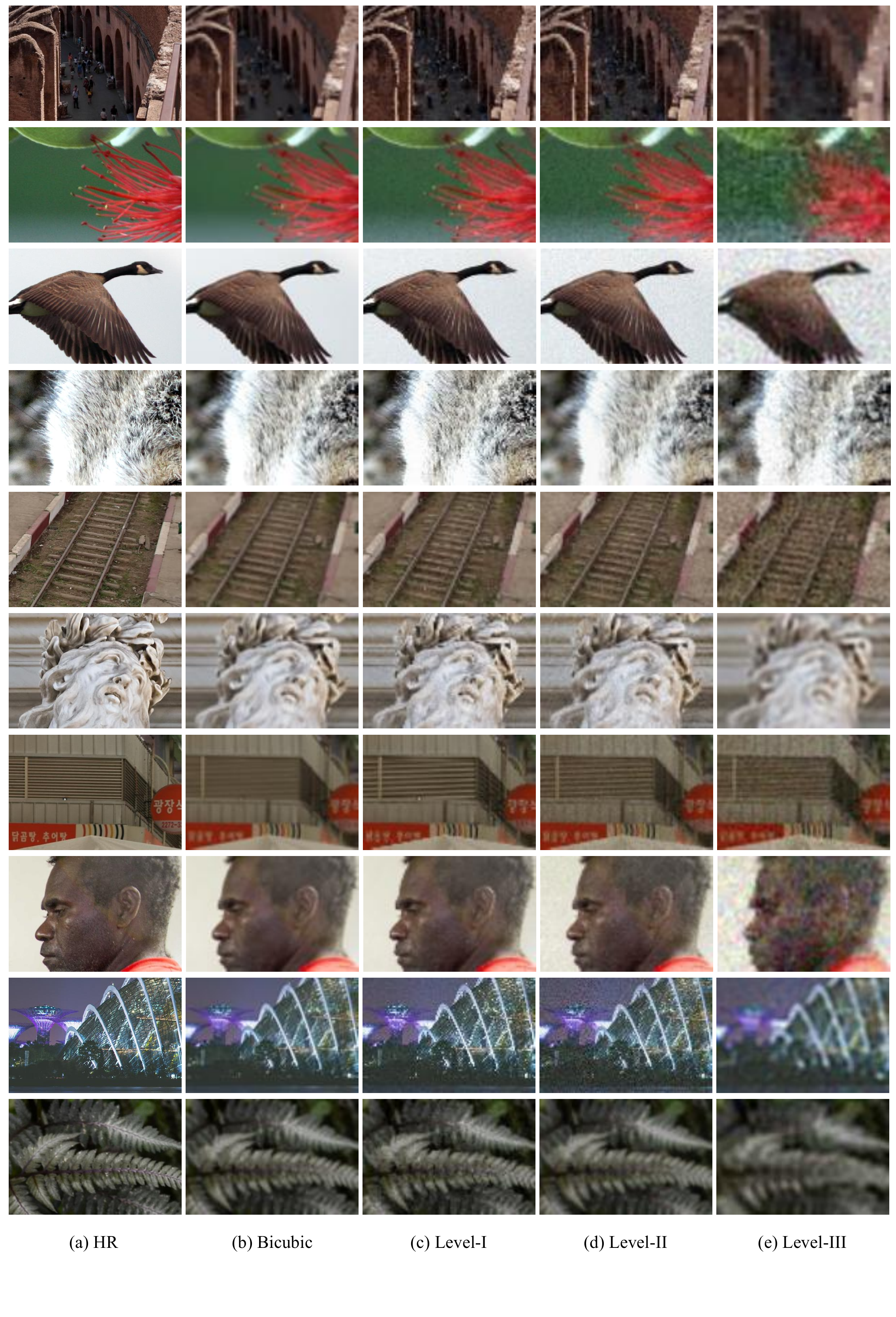}
		\vspace{-4.5em}
		\caption{\label{more_sample_image}More sample images with different levels of degradations in our constructed datasets, as well as the ground-truth HR images. Level-I, -II, and -III represent the samples whose degradations belong to $S_1$, $S_2$, and $S_3$, respectively.}\vspace{-1.5em}
	\end{figure*}

	In Fig.~\ref{more_sample_image}, we provide more sample images with different degradation levels in our datasets, as well as the ground-truth HR images. As can be seen from the figure, those images can cover a wide range of real-world degradations. The balanced sampling from the three levels during training improves the generalization capacity of our DASR to real-world images with different degradations.
	
	\subsection{More Qualitative Comparisons}
	\label{More_comparison}
	
\begin{figure*}[h]
		\centering
		\includegraphics[width=1\textwidth]{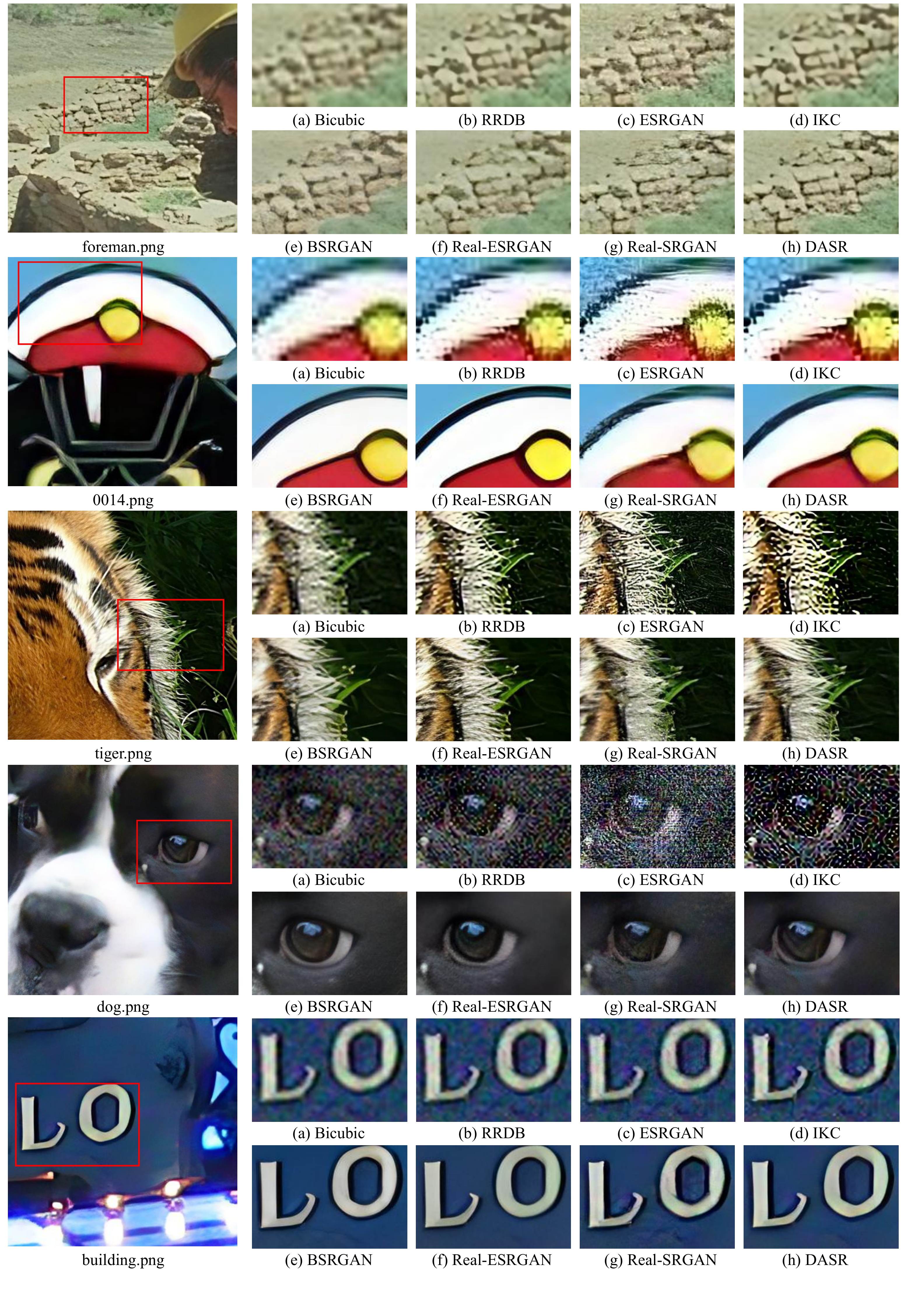}
		\vspace{-3.5em}
		\caption{\label{comparison_supp_real}More qualitative comparison of competing methods on \textbf{real-world} images. The results of (b-f) are generated by using the officially released models, while the output of (g) is obtained by re-training the SRResNet backbone with our proposed degradation model. Better zoom in for details.}\vspace{-1.5em}
	\end{figure*}

\begin{figure*}[h]
		\centering
		\includegraphics[width=1\textwidth]{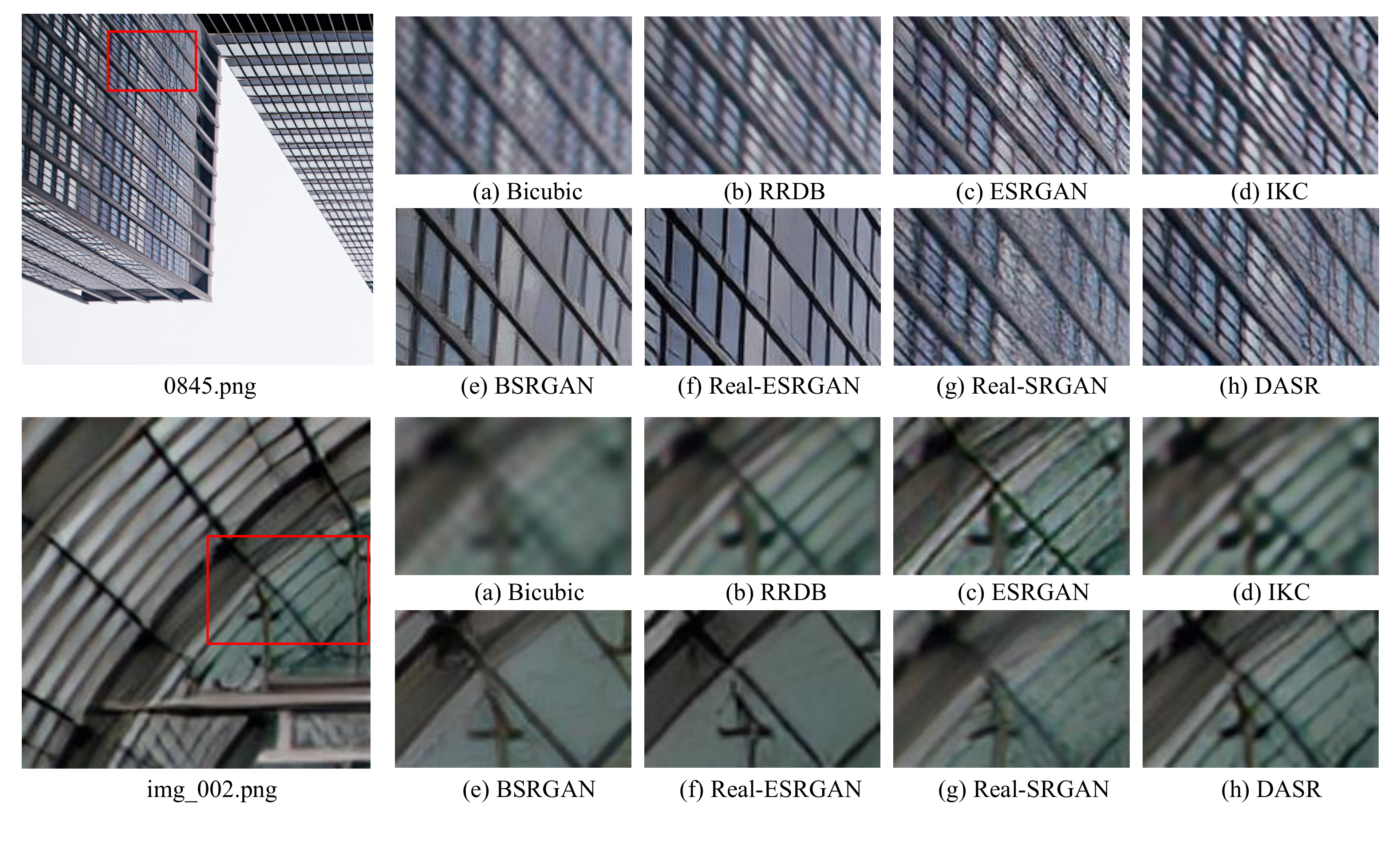}
		\vspace{-3.5em}
		\caption{\label{comparison_supp_div2k0}More qualitative comparison of competing methods on images with \textbf{bicubic} downsampling. The results of (b-f) are generated by using the officially released models, while the output of (g) is obtained by re-training the SRResNet backbone with our proposed degradation model. Better zoom in for details.}\vspace{-1.5em}
	\end{figure*}

\begin{figure*}[h]
		\centering
		\includegraphics[width=1\textwidth]{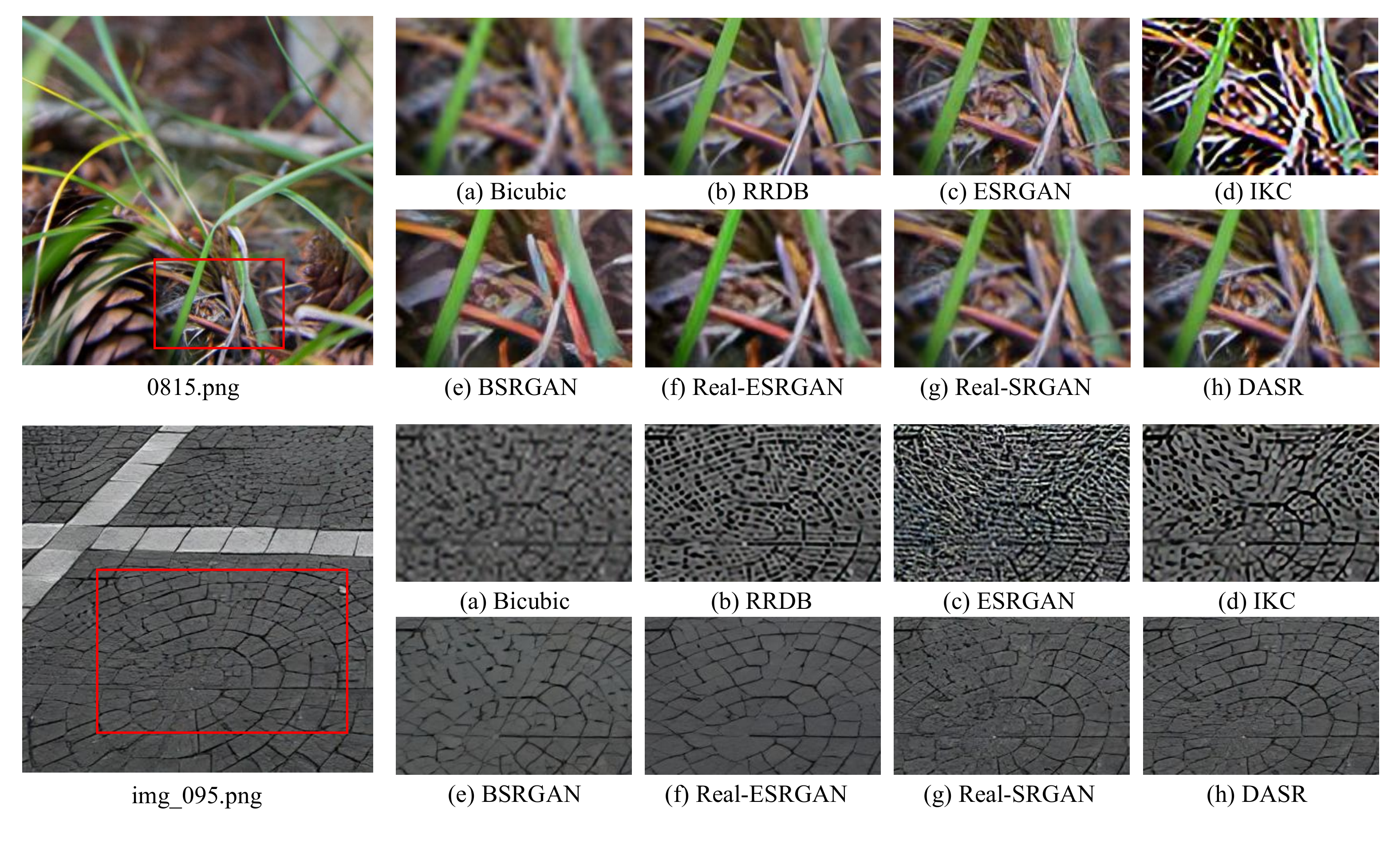}
		\vspace{-3.5em}
		\caption{\label{comparison_supp_div2k1}More qualitative comparison of competing methods on images with degradation of \textbf{Level-I}. The results of (b-f) are generated by using the officially released models, while the output of (g) is obtained by re-training the SRResNet backbone with our proposed degradation model. Better zoom in for details.}\vspace{-1.5em}
	\end{figure*}

\begin{figure*}[h]
		\centering
		\includegraphics[width=1\textwidth]{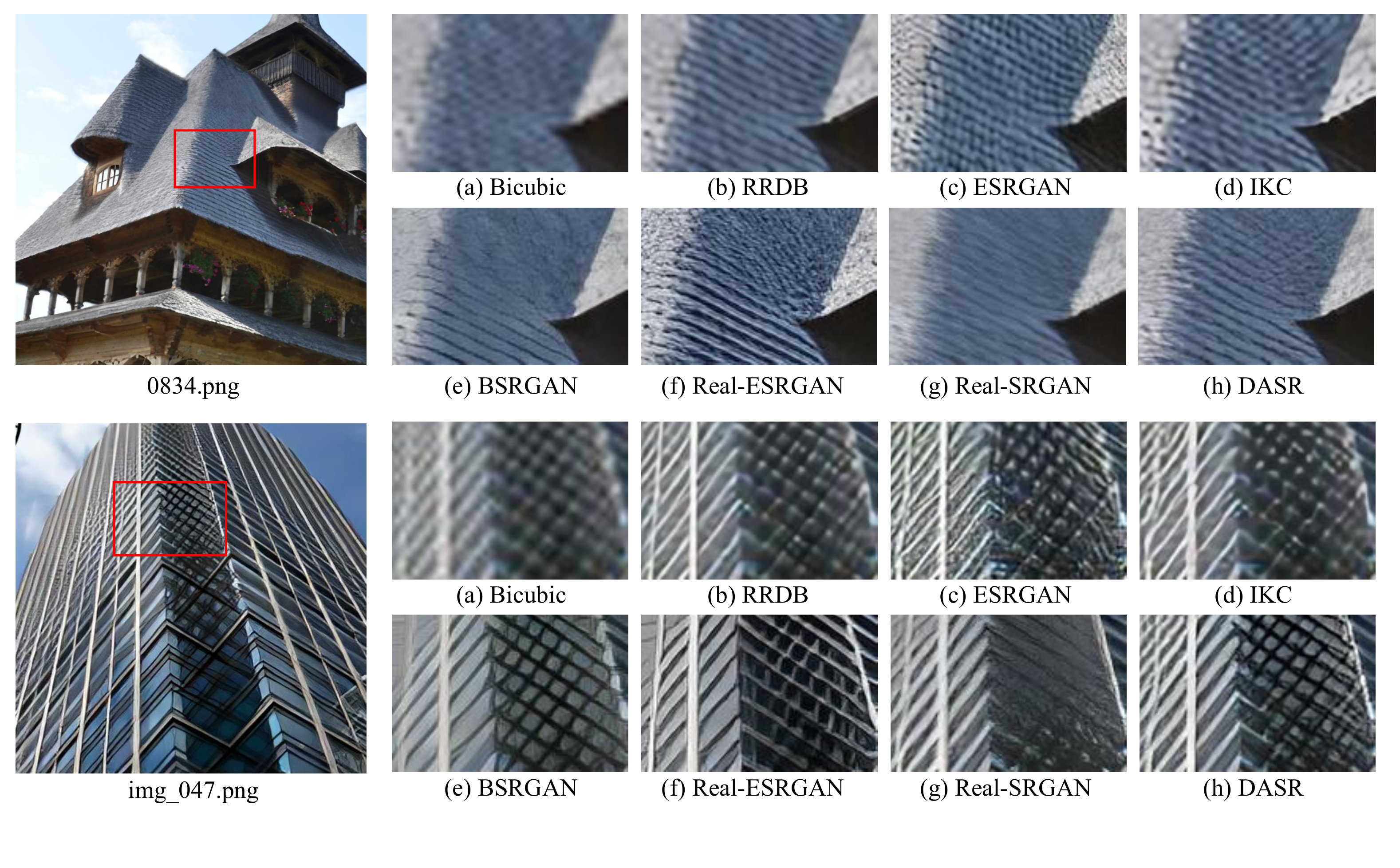}
		\vspace{-3.5em}
		\caption{\label{comparison_supp_div2k2}More qualitative comparison of competing methods on images with degradation of \textbf{Level-II}. The results of (b-f) are generated by using the officially released models, while the output of (g) is obtained by re-training the SRResNet backbone with our proposed degradation model. Better zoom in for details.}\vspace{-1.5em}
	\end{figure*}

\begin{figure*}[h]
		\centering
		\includegraphics[width=1\textwidth]{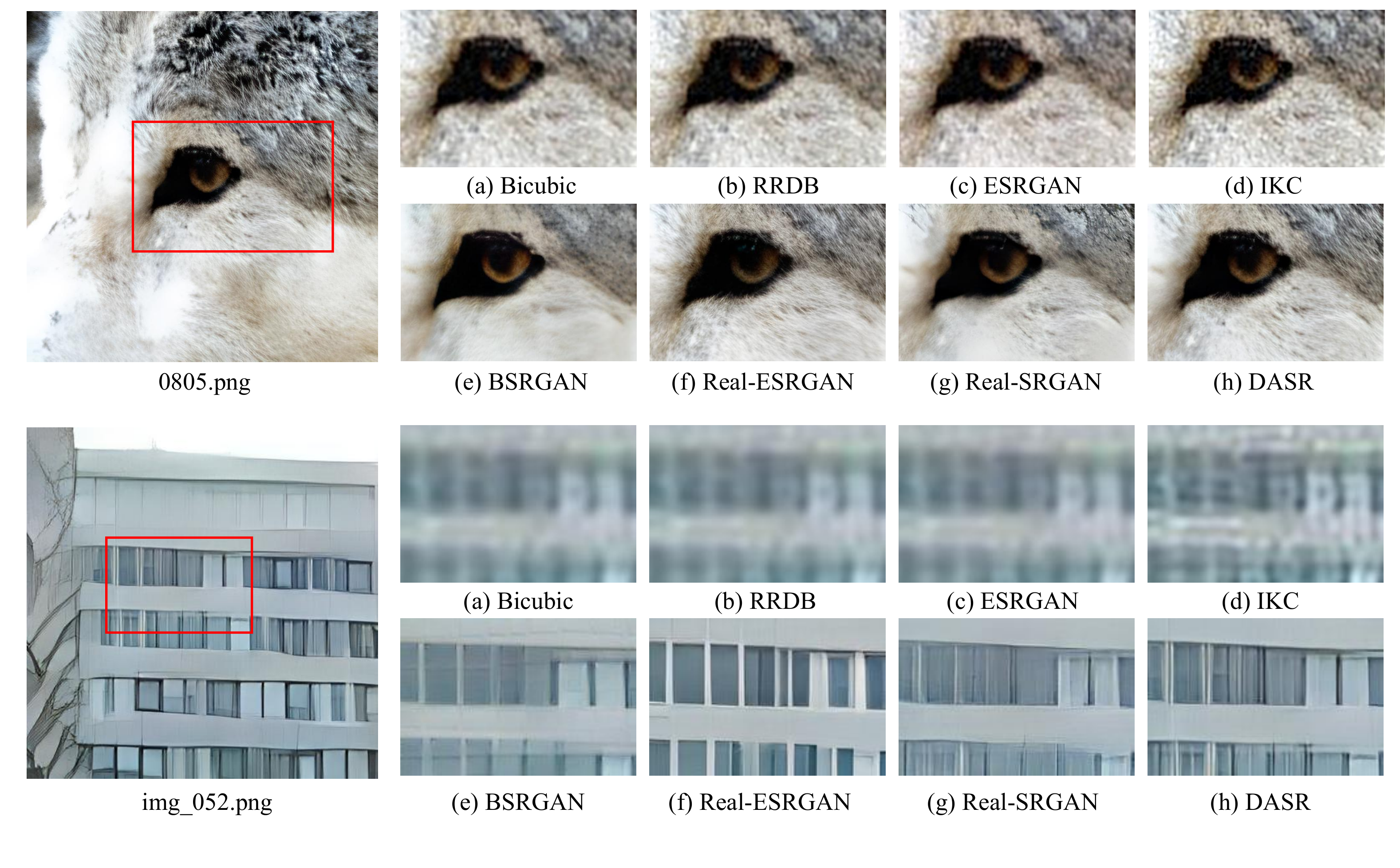}
		\vspace{-3.5em}
		\caption{\label{comparison_supp_div2k3}More qualitative comparison of competing methods on images with degradation of \textbf{Level-III}. The results of (b-f) are generated by using the officially released models, while the output of (g) is obtained by re-training the SRResNet backbone with our proposed degradation model. Better zoom in for details.}\vspace{-1.5em}
	\end{figure*}

	In Fig.~\ref{comparison_supp_real}, we provide more qualitative comparisons of competing methods on real-world images, while in Figs.~\ref{comparison_supp_div2k0},~\ref{comparison_supp_div2k1},~\ref{comparison_supp_div2k2} and~\ref{comparison_supp_div2k3}, we provide more qualitative comparisons of competing methods on datasets with bicubic, Level-I, Level-II and Level-III degradations, respectively. Our models are trained by using the images in DIV2K, Flickr2K, and OutdoorScene-Training datasets. To further validate the generalization capability of DASR to different image contents, the visual comparisons in Figs.~\ref{comparison_supp_div2k0},~\ref{comparison_supp_div2k1},~\ref{comparison_supp_div2k2} and~\ref{comparison_supp_div2k3} also include images from the Urban100 dataset by using the same degrading strategy as in our main paper. From those figures, consistent observations to our main paper can be made. Our DASR can generate more realistic structures and details on different degradations, benefiting from its degradation-adaptive strategy and the joint training and adaptive mixture of multiple experts. 
	
	\begin{table}[t]
		\caption{Detailed parameter settings of the degradation sub-spaces $[S_1, S_2, S_3]$. Here, `-' indicates that the operation is not activated and the corresponding value in $\bm{v}$ is padded with $0$; [`a', `b', `b'] denote the resize modes of [area, bilinear, bicubic]; [`G', `P'] denote the noise types of [Gaussian, Poisson]; $\omega_c$ is the cutoff frequency of the sinc kernel; R-J and J-R indicate the different operating orders of resizing and JPEG compression; $v_i$ denotes the $i^{th}$ value of $\bm{v}$.}
		\scriptsize
		\label{parameter}
		\vspace{-2em}
		\begin{center}
			\begin{tabular}{p{1cm}<{\centering}|p{1.3cm}<{\centering}|p{3cm}<{\centering}|p{1.5cm}<{\centering}p{1.2cm}<{\centering}|p{1.5cm}<{\centering}p{1.2cm}<{\centering}}
				\toprule
				\multirow{2}{*}{Level}&\multirow{2}{*}{Operation}&\multirow{2}{*}{Parameter}&\multicolumn{2}{c|}{Stage 1}&\multicolumn{2}{c}{Stage 2}\\
				&&&Range&$v_i$&Range&$v_i$\\
				\midrule
				\multirow{11}{*}{$S_1$}&\multirow{4}{*}{Blur}&kernel size $[2m+1]$&$m\in[3,10]$&$v_1$&-&-\\
				&&standard deviation $\sigma_1$&$ [0.2, 0.8] $&$v_2$&-&-\\
				&&standard deviation $\sigma_2$&$ [0.2, 0.8] $&$v_3$&-&-\\
				&&rotation degree $\theta$&$[-\pi, \pi]$&$v_4$&-&-\\
				\cmidrule(lr{0em}){2-7}
				&\multirow{3}{*}{Resize}&[up, down, keep]&$[0.1, 0.2, 0.7]$&-&-&-\\
				&&scale factor&$ [0.85, 1.2] $&$v_{11}$&-&-\\
				&&resize mode&[`a', `b', `b']&$v_{12}\sim v_{14}$&-&-\\
				\cmidrule(lr{0em}){2-7}
				&\multirow{4}{*}{Noise}&type&[`G', `P']&$v_{21}, v_{22}$&-&-\\
				&&sigma of Gaussian&$[1,10]$&$v_{19}$&-&-\\
				&&scale of Poisson&$[0.05, 0.5]$&$v_{19}$&-&-\\
				&&gray probability&$0.4$&$v_{20}$&-&-\\
				\cmidrule(lr{0em}){2-7}
				&\multirow{2}{*}{JPEG}&quality factor&$[90, 95]$&$v_{27}$&-&-\\
				&&mode of final resize&[`a', `b', `b']&$v_{31}\sim v_{33}$&-&-\\
				\midrule
				\multirow{16}{*}{$S_2$}&\multirow{6}{*}{Blur}&kernel size $[2m+1]$&$m\in[3,10]$&$v_1$&-&-\\
				&&standard deviation $\sigma_1$&$[0.2, 1.5]$&$v_2$&-&-\\
				&&standard deviation $\sigma_2$&$[0.2, 1.5]$&$v_3$&-&-\\
				&&rotation degree $\theta$&$[-\pi, \pi]$&$v_4$&-&-\\
				\cmidrule(lr{0em}){2-7}
				&\multirow{3}{*}{Resize}&[up, down, keep]&$[0.3, 0.4, 0.3]$&-&-&-\\
				&&scale factor&$ [0.5, 1.2] $&$v_{11}$&-&-\\
				&&resize mode&[`a', `b', `b']&$v_{12}\sim v_{14}$&-&-\\
				\cmidrule(lr{0em}){2-7}
				&\multirow{4}{*}{Noise}&type&[`G', `P']&$v_{21}, v_{22}$&-&-\\
				&&sigma of Gaussian&$[1,20]$&$v_{19}$&-&-\\
				&&scale of Poisson&$[0.05, 1.5]$&$v_{19}$&-&-\\
				&&gray probability&$0.4$&$v_{20}$&-&-\\
				\cmidrule(lr{0em}){2-7}
				&\multirow{2}{*}{JPEG}&quality factor&$[50, 95]$&$v_{27}$&-&-\\
				&&mode of final resize&[`a', `b', `b']&$v_{31}\sim v_{33}$&-&-\\
				\midrule
				\multirow{16}{*}{$S_3$}&\multirow{6}{*}{Blur}&kernel size $[2m+1]$&$m\in[3,10]$&$v_1$&$m\in[3,10]$&$v_5$\\
				&&standard deviation $\sigma_1$&$[0.2, 3]$&$v_2$&$[0.2, 1.5]$&$v_6$\\
				&&standard deviation $\sigma_2$&$[0.2, 3]$&$v_3$&$[0.2, 1.5]$&$v_7$\\
				&&rotation degree $\theta$&$[-\pi, \pi]$&$v_4$&$[-\pi, \pi]$&$v_8$\\
				&&sinc kernel size $[2m+1]$&-&-&$m\in[3,10]$&$v_9$\\
				&&$\omega_c$ of sinc kernel&-&-&$[\pi/3, \pi]$&$v_{10}$\\
				\cmidrule(lr{0em}){2-7}
				&\multirow{3}{*}{Resize}&[up, down, keep]&$[0.2, 0.7, 0.1]$&-&$[0.3, 0.4, 0.3]$&-\\
				&&scale factor&$ [0.15, 1.5] $&$v_{11}$&$ [0.3, 1.2] $&$v_{15}$\\
				&&resize mode&[`a', `b', `b']&$v_{12}\sim v_{14}$&[`a', `b', `b']&$v_{16}\sim v_{18}$\\
				\cmidrule(lr{0em}){2-7}
				&\multirow{4}{*}{Noise}&type&[`G', `P']&$v_{21}, v_{22}$&[`G', `P']&$v_{25}, v_{26}$\\
				&&sigma of Gaussian&$[1,30]$&$v_{19}$&$[1, 25]$&$v_{23}$\\
				&&scale of Poisson&$[0.05, 3]$&$v_{19}$&$[0.05, 2.5]$&$v_{23}$\\
				&&gray probability&$0.4$&$v_{20}$&$0.4$&$v_{24}$\\
				\cmidrule(lr{0em}){2-7}
				&\multirow{3}{*}{JPEG}&quality factor&$[30, 95]$&$v_{27}$&$[30, 95]$&$v_{28}$\\
				&&operating order&-&-&R-J or J-R&$v_{29}, v_{30}$\\
				&&mode of final resize&-&-&[`a', `b', `b']&$v_{31}\sim v_{33}$\\
				\bottomrule
			\end{tabular}
		\end{center}\vspace{-2em}
	\end{table}

	
	\clearpage
	\bibliographystyle{splncs04}
	\bibliography{bib_DASR}

\begin{thebibliography}{10}

\bibitem{johnson2016perceptual}
Justin Johnson, Alexandre Alahi, and Li~Fei-Fei.
\newblock Perceptual losses for real-time style transfer and super-resolution.
\newblock In {\em ECCV}, 2016.

\bibitem{wang2018esrgan}
Xintao Wang, Ke~Yu, Shixiang Wu, Jinjin Gu, Yihao Liu, Chao Dong, Yu~Qiao, and
  Chen Change~Loy.
\newblock {ESRGAN}: Enhanced super-resolution generative adversarial networks.
\newblock In {\em ECCVW}, 2018.

\bibitem{zhang2018image}
Yulun Zhang, Kunpeng Li, Kai Li, Lichen Wang, Bineng Zhong, and Yun Fu.
\newblock Image super-resolution using very deep residual channel attention
  networks.
\newblock In {\em ECCV}, 2018.

\bibitem{ma2020structure}
Cheng Ma, Yongming Rao, Yean Cheng, Ce~Chen, Jiwen Lu, and Jie Zhou.
\newblock Structure-preserving super resolution with gradient guidance.
\newblock In {\em CVPR}, 2020.

\bibitem{sun2010gradient}
Jian Sun, Zongben Xu, and Heung-Yeung Shum.
\newblock Gradient profile prior and its applications in image super-resolution
  and enhancement.
\newblock {\em IEEE Transactions on Image Processing}, 20(6):1529--1542, 2010.

\bibitem{dong2014learning}
Chao Dong, Chen~Change Loy, Kaiming He, and Xiaoou Tang.
\newblock Learning a deep convolutional network for image super-resolution.
\newblock In {\em ECCV}, 2014.

\bibitem{sajjadi2017enhancenet}
Mehdi~SM Sajjadi, Bernhard Scholkopf, and Michael Hirsch.
\newblock Enhancenet: Single image super-resolution through automated texture
  synthesis.
\newblock In {\em ICCV}, 2017.

\bibitem{kim2016deeply}
Jiwon Kim, Jung~Kwon Lee, and Kyoung~Mu Lee.
\newblock Deeply-recursive convolutional network for image super-resolution.
\newblock In {\em CVPR}, 2016.

\bibitem{soh2019natural}
Jae~Woong Soh, Gu~Yong Park, Junho Jo, and Nam~Ik Cho.
\newblock Natural and realistic single image super-resolution with explicit
  natural manifold discrimination.
\newblock In {\em CVPR}, 2019.

\bibitem{zhang2018residual}
Yulun Zhang, Yapeng Tian, Yu~Kong, Bineng Zhong, and Yun Fu.
\newblock Residual dense network for image super-resolution.
\newblock In {\em CVPR}, 2018.

\bibitem{jo2021tackling}
Younghyun Jo, Seoung~Wug Oh, Peter Vajda, and Seon~Joo Kim.
\newblock Tackling the ill-posedness of super-resolution through adaptive
  target generation.
\newblock In {\em CVPR}, 2021.

\bibitem{liu2021blind}
Anran Liu, Yihao Liu, Jinjin Gu, Yu~Qiao, and Chao Dong.
\newblock Blind image super-resolution: A survey and beyond.
\newblock {\em arXiv preprint arXiv:2107.03055}, 2021.

\bibitem{zhang2021designing}
Kai Zhang, Jingyun Liang, Luc Van~Gool, and Radu Timofte.
\newblock Designing a practical degradation model for deep blind image
  super-resolution.
\newblock In {\em ICCV}, 2021.

\bibitem{zhang2018learning}
Kai Zhang, Wangmeng Zuo, and Lei Zhang.
\newblock Learning a single convolutional super-resolution network for multiple
  degradations.
\newblock In {\em CVPR}, 2018.

\bibitem{luo2020unfolding}
Zhengxiong Luo, Yan Huang, Shang Li, Liang Wang, and Tieniu Tan.
\newblock Unfolding the alternating optimization for blind super resolution.
\newblock In {\em NeurIPS}, 2020.

\bibitem{gu2019blind}
Jinjin Gu, Hannan Lu, Wangmeng Zuo, and Chao Dong.
\newblock Blind super-resolution with iterative kernel correction.
\newblock In {\em CVPR}, 2019.

\bibitem{zhou2019kernel}
Ruofan Zhou and Sabine Susstrunk.
\newblock Kernel modeling super-resolution on real low-resolution images.
\newblock In {\em ICCV}, 2019.

\bibitem{ignatov2017dslr}
Andrey Ignatov, Nikolay Kobyshev, Radu Timofte, Kenneth Vanhoey, and Luc
  Van~Gool.
\newblock {DSLR}-quality photos on mobile devices with deep convolutional
  networks.
\newblock In {\em ICCV}, 2017.

\bibitem{cai2019toward}
Jianrui Cai, Hui Zeng, Hongwei Yong, Zisheng Cao, and Lei Zhang.
\newblock Toward real-world single image super-resolution: A new benchmark and
  a new model.
\newblock In {\em ICCV}, 2019.

\bibitem{wei2020component}
Pengxu Wei, Ziwei Xie, Hannan Lu, Zongyuan Zhan, Qixiang Ye, Wangmeng Zuo, and
  Liang Lin.
\newblock Component divide-and-conquer for real-world image super-resolution.
\newblock In {\em ECCV}, 2020.

\bibitem{lugmayr2020ntire}
Andreas Lugmayr, Martin Danelljan, and Radu Timofte.
\newblock {NTIRE} 2020 challenge on real-world image super-resolution: Methods
  and results.
\newblock In {\em CVPRW}, 2020.

\bibitem{lugmayr2019aim}
Andreas Lugmayr, Martin Danelljan, Radu Timofte, Manuel Fritsche, Shuhang Gu,
  Kuldeep Purohit, Praveen Kandula, Maitreya Suin, AN~Rajagoapalan, Nam~Hyung
  Joon, et~al.
\newblock {AIM} 2019 challenge on real-world image super-resolution: Methods
  and results.
\newblock In {\em ICCVW}, 2019.

\bibitem{fritsche2019frequency}
Manuel Fritsche, Shuhang Gu, and Radu Timofte.
\newblock Frequency separation for real-world super-resolution.
\newblock In {\em ICCVW}, 2019.

\bibitem{lugmayr2019unsupervised}
Andreas Lugmayr, Martin Danelljan, and Radu Timofte.
\newblock Unsupervised learning for real-world super-resolution.
\newblock In {\em ICCVW}, 2019.

\bibitem{ji2020real}
Xiaozhong Ji, Yun Cao, Ying Tai, Chengjie Wang, Jilin Li, and Feiyue Huang.
\newblock Real-world super-resolution via kernel estimation and noise
  injection.
\newblock In {\em CVPRW}, 2020.

\bibitem{ren2020real}
Haoyu Ren, Amin Kheradmand, Mostafa El-Khamy, Shuangquan Wang, Dongwoon Bai,
  and Jungwon Lee.
\newblock Real-world super-resolution using generative adversarial networks.
\newblock In {\em CVPRW}, 2020.

\bibitem{bulat2018learn}
Adrian Bulat, Jing Yang, and Georgios Tzimiropoulos.
\newblock To learn image super-resolution, use a {GAN} to learn how to do image
  degradation first.
\newblock In {\em ECCV}, 2018.

\bibitem{wei2021unsupervised}
Yunxuan Wei, Shuhang Gu, Yawei Li, Radu Timofte, Longcun Jin, and Hengjie Song.
\newblock Unsupervised real-world image super resolution via domain-distance
  aware training.
\newblock In {\em CVPR}, 2021.

\bibitem{wang2021realesrgan}
Xintao Wang, Liangbin Xie, Chao Dong, and Ying Shan.
\newblock {Real-ESRGAN}: Training real-world blind super-resolution with pure
  synthetic data.
\newblock In {\em ICCVW}, 2021.

\bibitem{dong2016accelerating}
Chao Dong, Change~Loy Chen, and Tang Xiaoou.
\newblock Accelerating the super-resolution convolutional neural network.
\newblock In {\em ECCV}, 2016.

\bibitem{zhang2021edge}
Xindong Zhang, Hui Zeng, and Lei Zhang.
\newblock Edge-oriented convolution block for real-time super resolution on
  mobile devices.
\newblock In {\em ACM Multimedia}, 2021.

\bibitem{ahn2018fast}
Namhyuk Ahn, Byungkon Kang, and Kyung-Ah Sohn.
\newblock Fast, accurate, and lightweight super-resolution with cascading
  residual network.
\newblock In {\em ECCV}, 2018.

\bibitem{yang2019lightweight}
Wenming Yang, Wei Wang, Xuechen Zhang, Shuifa Sun, and Qingmin Liao.
\newblock Lightweight feature fusion network for single image super-resolution.
\newblock {\em IEEE Signal Processing Letters}, 26(4):538--542, 2019.

\bibitem{song2021addersr}
Dehua Song, Yunhe Wang, Hanting Chen, Chang Xu, Chunjing Xu, and DaCheng Tao.
\newblock Addersr: Towards energy efficient image super-resolution.
\newblock In {\em CVPR}, 2021.

\bibitem{zhang2019zoom}
Xuaner Zhang, Qifeng Chen, Ren Ng, and Vladlen Koltun.
\newblock Zoom to learn, learn to zoom.
\newblock In {\em CVPR}, 2019.

\bibitem{liang2021swinir}
Jingyun Liang, Jiezhang Cao, Guolei Sun, Kai Zhang, Luc Van~Gool, and Radu
  Timofte.
\newblock {SwinIR}: Image restoration using swin transformer.
\newblock In {\em ICCVW}, 2021.

\bibitem{liu2021swin}
Ze~Liu, Yutong Lin, Yue Cao, Han Hu, Yixuan Wei, Zheng Zhang, Stephen Lin, and
  Baining Guo.
\newblock Swin transformer: Hierarchical vision transformer using shifted
  windows.
\newblock In {\em ICCV}, 2021.

\bibitem{ledig2017photo}
Christian Ledig, Lucas Theis, Ferenc Husz{\'a}r, Jose Caballero, Andrew
  Cunningham, Alejandro Acosta, Andrew Aitken, Alykhan Tejani, Johannes Totz,
  Zehan Wang, et~al.
\newblock Photo-realistic single image super-resolution using a generative
  adversarial network.
\newblock In {\em CVPR}, 2017.

\bibitem{wang2018recovering}
Xintao Wang, Ke~Yu, Chao Dong, and Chen~Change Loy.
\newblock Recovering realistic texture in image super-resolution by deep
  spatial feature transform.
\newblock In {\em CVPR}, 2018.

\bibitem{fuoli2021fourier}
Dario Fuoli, Luc Van~Gool, and Radu Timofte.
\newblock Fourier space losses for efficient perceptual image super-resolution.
\newblock In {\em ICCV}, 2021.

\bibitem{xu2020unified}
Yu-Syuan Xu, Shou-Yao~Roy Tseng, Yu~Tseng, Hsien-Kai Kuo, and Yi-Min Tsai.
\newblock Unified dynamic convolutional network for super-resolution with
  variational degradations.
\newblock In {\em CVPR}, 2020.

\bibitem{zhang2020deep}
Kai Zhang, Luc~Van Gool, and Radu Timofte.
\newblock Deep unfolding network for image super-resolution.
\newblock In {\em CVPR}, 2020.

\bibitem{zhang2019deep}
Kai Zhang, Wangmeng Zuo, and Lei Zhang.
\newblock Deep plug-and-play super-resolution for arbitrary blur kernels.
\newblock In {\em CVPR}, 2019.

\bibitem{wang2021unsupervised}
Longguang Wang, Yingqian Wang, Xiaoyu Dong, Qingyu Xu, Jungang Yang, Wei An,
  and Yulan Guo.
\newblock Unsupervised degradation representation learning for blind
  super-resolution.
\newblock In {\em CVPR}, 2021.

\bibitem{hui2021learning}
Zheng Hui, Jie Li, Xiumei Wang, and Xinbo Gao.
\newblock Learning the non-differentiable optimization for blind
  super-resolution.
\newblock In {\em CVPR}, 2021.

\bibitem{liu2020learning}
Pengju Liu, Hongzhi Zhang, Yue Cao, Shigang Liu, Dongwei Ren, and Wangmeng Zuo.
\newblock Learning cascaded convolutional networks for blind single image
  super-resolution.
\newblock {\em Neurocomputing}, 417:371--383, 2020.

\bibitem{maeda2020unpaired}
Shunta Maeda.
\newblock Unpaired image super-resolution using pseudo-supervision.
\newblock In {\em CVPR}, 2020.

\bibitem{yuan2018unsupervised}
Yuan Yuan, Siyuan Liu, Jiawei Zhang, Yongbing Zhang, Chao Dong, and Liang Lin.
\newblock Unsupervised image super-resolution using cycle-in-cycle generative
  adversarial networks.
\newblock In {\em CVPRW}, 2018.

\bibitem{goodfellow2014generative}
Ian Goodfellow, Jean Pouget-Abadie, Mehdi Mirza, Bing Xu, David Warde-Farley,
  Sherjil Ozair, Aaron Courville, and Yoshua Bengio.
\newblock Generative adversarial nets.
\newblock {\em NeurIPS}, 2014.

\bibitem{cornillere2019blind}
Victor Cornillere, Abdelaziz Djelouah, Wang Yifan, Olga Sorkine-Hornung, and
  Christopher Schroers.
\newblock Blind image super-resolution with spatially variant degradations.
\newblock {\em ACM Transactions on Graphics (TOG)}, 38(6):1--13, 2019.

\bibitem{kim2021koalanet}
Soo~Ye Kim, Hyeonjun Sim, and Munchurl Kim.
\newblock {KOALAnet}: Blind super-resolution using kernel-oriented adaptive
  local adjustment.
\newblock In {\em CVPR}, 2021.

\bibitem{jacobs1991adaptive}
Robert~A Jacobs, Michael~I Jordan, Steven~J Nowlan, and Geoffrey~E Hinton.
\newblock Adaptive mixtures of local experts.
\newblock {\em Neural computation}, 3(1):79--87, 1991.

\bibitem{jordan1994hierarchical}
Michael~I Jordan and Robert~A Jacobs.
\newblock Hierarchical mixtures of experts and the em algorithm.
\newblock {\em Neural computation}, 6(2):181--214, 1994.

\bibitem{gross2017hard}
Sam Gross, Marc'Aurelio Ranzato, and Arthur Szlam.
\newblock Hard mixtures of experts for large scale weakly supervised vision.
\newblock In {\em CVPR}, 2017.

\bibitem{aljundi2017expert}
Rahaf Aljundi, Punarjay Chakravarty, and Tinne Tuytelaars.
\newblock Expert gate: Lifelong learning with a network of experts.
\newblock In {\em CVPR}, 2017.

\bibitem{maeda2020fast}
Shunta Maeda.
\newblock Fast and flexible image blind denoising via competition of experts.
\newblock In {\em CVPRW}, 2020.

\bibitem{wang2020blind}
Yifan Wang, Lijun Wang, Hongyu Wang, Peihua Li, and Huchuan Lu.
\newblock Blind single image super-resolution with a mixture of deep networks.
\newblock {\em Pattern Recognition}, 102:107169, 2020.

\bibitem{chen2020dynamic}
Yinpeng Chen, Xiyang Dai, Mengchen Liu, Dongdong Chen, Lu~Yuan, and Zicheng
  Liu.
\newblock Dynamic convolution: Attention over convolution kernels.
\newblock In {\em CVPR}, 2020.

\bibitem{li2021revisiting}
Yunsheng Li, Yinpeng Chen, Xiyang Dai, Mengchen Liu, Dongdong Chen, Ye~Yu,
  Lu~Yuan, Zicheng Liu, Mei Chen, and Nuno Vasconcelos.
\newblock Revisiting dynamic convolution via matrix decomposition.
\newblock In {\em ICLR}, 2021.

\bibitem{li2021omni}
Chao Li, Aojun Zhou, and Anbang Yao.
\newblock Omni-dimensional dynamic convolution.
\newblock In {\em ICLR}, 2021.

\bibitem{yang2019condconv}
Brandon Yang, Gabriel Bender, Quoc~V Le, and Jiquan Ngiam.
\newblock Condconv: Conditionally parameterized convolutions for efficient
  inference.
\newblock {\em NeurIPS}, 2019.

\bibitem{lim2017enhanced}
Bee Lim, Sanghyun Son, Heewon Kim, Seungjun Nah, and Kyoung Mu~Lee.
\newblock Enhanced deep residual networks for single image super-resolution.
\newblock In {\em CVPRW}, 2017.

\bibitem{kingma2014adam}
Diederik~P Kingma and Jimmy Ba.
\newblock Adam: A method for stochastic optimization.
\newblock {\em arXiv preprint arXiv:1412.6980}, 2014.

\bibitem{zhang2020aim}
Kai Zhang, Martin Danelljan, Yawei Li, Radu Timofte, et~al.
\newblock Aim 2020 challenge on efficient super-resolution: Methods and
  results.
\newblock In {\em ECCVW}, 2020.

\bibitem{zhang2019aim}
Kai Zhang, Shuhang Gu, Radu Timofte, et~al.
\newblock Aim 2019 challenge on constrained super-resolution: Methods and
  results.
\newblock In {\em ICCVW}, 2019.

\bibitem{emad2022moesr}
Mohammad Emad, Maurice Peemen, and Henk Corporaal.
\newblock {MoESR}: Blind super-resolution using kernel-aware mixture of
  experts.
\newblock In {\em WACV}, 2022.

\bibitem{simonyan2014very}
Karen Simonyan and Andrew Zisserman.
\newblock Very deep convolutional networks for large-scale image recognition.
\newblock In {\em ICLR}, 2015.

\end{thebibliography}
\end{document}